\documentclass[conference]{IEEEtran}
\usepackage{times}
\usepackage{graphics} 
\usepackage{epsfig} 
\usepackage{mathptmx} 
\usepackage{times} 
\usepackage{amsmath} 
\usepackage{amssymb}  
\usepackage{color}
\usepackage{url}
\usepackage{caption}
\usepackage{subcaption}
\usepackage{array}
\usepackage{multirow}
\usepackage{algorithm}
\usepackage{algpseudocode}
\newcolumntype{P}[1]{>{\centering\arraybackslash}p{#1}}
\newcolumntype{M}[1]{>{\centering\arraybackslash}m{#1}}
\usepackage{comment}
\usepackage{siunitx}

\usepackage{xargs}
\usepackage{xcolor}  
\usepackage[colorinlistoftodos,prependcaption,textsize=tiny]{todonotes}
\newcommandx{\Sen}[2][1=]{\todo[inline,linecolor=red,backgroundcolor=red!25,bordercolor=red,#1]{#2}}

\usepackage[numbers]{natbib}
\usepackage{multicol}

\begin{document}
\graphicspath{ {./figures/} }

\title{{\LARGE CURL: Continuous, Ultra-compact Representation for LiDAR}}




%
\author{
\authorblockN{Kaicheng Zhang,
Ziyang Hong,
Shida Xu,
 and
Sen Wang\IEEEauthorrefmark{1}} \

\authorblockA{Perception and Robotics Group, Institute of Signals, Sensors and Systems, Heriot-Watt University, UK\\
Email: \{kz13, zh9, sx2000, s.wang\}@hw.ac.uk \qquad *Corresponding author}
}

\maketitle

\begin{abstract}
Increasing the density of the 3D LiDAR point cloud is appealing for many applications in robotics. However, high-density LiDAR sensors are usually costly and still limited to a level of coverage per scan (e.g., 128 channels). Meanwhile, denser point cloud scans and maps mean larger volumes to store and longer times to transmit. Existing works focus on either improving point cloud density or compressing its size. This paper aims to design a novel 3D point cloud representation that can continuously increase point cloud density while reducing its storage and transmitting size. The pipeline of the proposed Continuous, Ultra-compact Representation of LiDAR (\textit{CURL}) includes four main steps: meshing, upsampling, encoding, and continuous reconstruction. It is capable of transforming a 3D LiDAR scan or map into a compact spherical harmonics representation which can be used or transmitted in low latency to continuously reconstruct a much denser 3D point cloud. Extensive experiments on four public datasets, covering college gardens, city streets, and indoor rooms, demonstrate that much denser 3D point clouds can be accurately reconstructed using the proposed CURL representation while achieving up to $80\%$ storage space-saving. We open-source the CURL codes for the community.
\end{abstract}

\IEEEpeerreviewmaketitle

\section{Introduction}

Light Detection and Ranging (LiDAR) sensing is key for a wide variety of robotic applications and mobile autonomy, e.g., self-driving vehicles, because it usually provides accurate and compelling ranging scans in the format of the 3D point cloud. However, a 3D point cloud scan provided by most low-cost, moderate-size LiDAR sensors is often sparse, particularly in large-scale outdoor scenarios. This could cause potential problems, e.g., miss detections of distant objects. Meanwhile, 3D point cloud scans are enormous in terms of data storage and memory usage. It can consume a large volume of storage to save 3D point clouds. More importantly, for remote and cloud applications that are increasingly in high demand, transmitting raw 3D point clouds can quickly saturate communication bandwidth, causing undesired transmitting latency or data package losses.

Some existing works, mainly based on deep learning, have achieved impressive performance on increasing the density of a point cloud \cite{yu2018pu,shan2020simulation}. Since they focus on generating a denser point cloud, the point cloud size would inevitably be maintained or increased for storage and transfer. On the other hand, some techniques are designed to compress point cloud \cite{cao20193d}, e.g., the video compression inspired MPEG standard for point cloud compression \cite{schwarz2018emerging}. However, they only target point cloud data compression and do not provide strategies for improving point cloud density.


This paper proposes a novel 3D point cloud representation that aims to address these challenges. It not only continuously and adaptively adjusts a point cloud's density on demand, but also models it in a compact format that can be efficiently stored and/or transmitted through a low communication bandwidth.
Our main contributions include:
\begin{enumerate}
  \item a novel 3D point cloud representation pipeline, named \textit{CURL}, that is designed to encode a 3D LiDAR point cloud in a continuous and ultra-compact format, i.e., an upsampled spherical harmonics representation for continuous point cloud reconstruction.
  \item an efficient meshing method and an upsampling technique developed to enhance density and accuracy of reconstructed 3D LiDAR point clouds in robotic scenarios, addressing the sparsity problem of raw 3D LiDAR scans, particularly in large-scale outdoor scenarios.
\end{enumerate}
The experiment evaluation on four public datasets shows that CURL is able to reduce the storage spaces of point clouds by up to $80\%$ while recovering denser 3D reconstructions.
To the best of our knowledge, this is the first framework that is capable of compressing a 3D point cloud into this level of storage efficiency while being able to (continuously) increase point cloud density.
Note that this work focuses on the widely used $360^\circ$ Field-of-View (FoV) LiDAR scanners, like Velodyne and Ouster ones, instead of  rectangular FoV sensors, e.g., Livox, although it is straightforward to adapt CURL for different scanning mechanisms.
The codes of CURL are released to promote this area in the community\footnote{\url{https://github.com/perception-and-robotics-group/CURL.git}}.

\begin{figure}
\includegraphics[width=\linewidth]{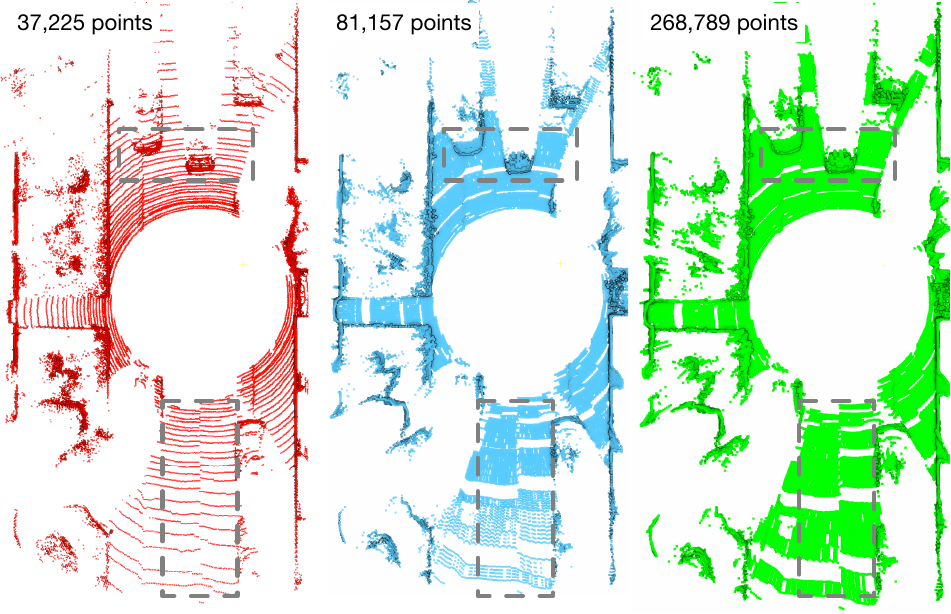}
\caption{Left to Right: original point cloud ({\color{red}red}), point clouds reconstructed using a same CURL with $2$ times ({\color{blue}blue}) and $7$ times ({\color{green}green}) density increases. The CURL of this point cloud is only $16\%$ of the original point cloud size.}
\label{fig:comparsion}
\end{figure}

\begin{figure}[h]
\includegraphics[width=\linewidth]{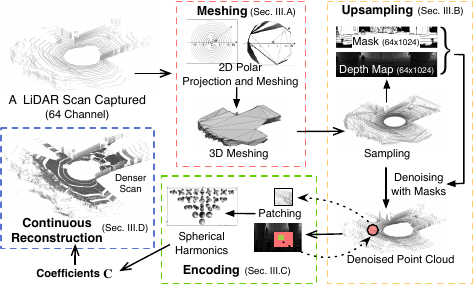}
\caption{Proposed CURL framework.}
\label{fig:pipeline}
\end{figure}

\section{Related Work}
A 3D point cloud contains a set of 3D points carrying both geometric and attribute information, e.g., intensity properties, which is the raw data often received from an off-the-shelf LiDAR sensor. Most existing works focus on designing descriptors of point clouds for various perception tasks, e.g., place recognition, object detection, and matching. These perception tasks heavily rely on high-resolution 3D point clouds, which directly and precisely describe the geometries of scenes and objects.
However, high-resolution point clouds take a large amount of storage to store and bandwidth to transmit. On the other hand, if the point cloud is not dense enough, it may hinder the performance of these perception tasks. Therefore, both upsampling and compression are essential for robotic applications using 3D point clouds.
\subsection{Point Cloud Upsampling}
Point cloud upsampling aims to increase the density of a point cloud. It is a challenging task because some geometric structures may not be fully encapsulated in sparse point clouds.
\subsubsection{Traditional approaches}
Upsampling is achieved in \cite{1175093} by adding points on the vertices of Voronoi diagrams on moving least squares surface. Surface reconstruction and point samples are conducted by calculating the characteristic functions with Fourier coefficients \cite{kazhdan2005reconstruction}. In \cite{wu2015deep}, the point density is increased by sinking the inner points to form a meso-skeleton and moving outer points along the surface to complete missing areas. \cite{pauly2001spectral} uses a set of window Fourier to create a spectral decomposition of the object model, which could be used to produce a dense reconstruction. The experiments of the above methods are all tested on enclosed point cloud of object models that are usually in small scales with little noise, different from outdoor environments where robots operate.
\subsubsection{Deep learning approaches}
A deep learning method that generates up-sampling results by learning multi-level features per point is proposed in \cite{yu2018pu}. \cite{yu2018ec} achieves the first deep learning-based edge-aware technique to facilitate the consolidation of point clouds. Instead of learning from patches, \cite{zhang2019data} learns from entire objects. \cite{qian2020pugeo} upsamples point clouds by projecting local surfaces into 2D, then using 3D linear transformation to recover 3D information. To achieve better multi-objective nature of the task, \cite{li2021point} designs one network cascaded by two networks. Previously, deep learning-based upsampling methods were all for relatively small objects. \cite{shan2020simulation} first proposes a technique that upsamples large-scale LiDAR point clouds from 16 to 64 channels, which will be compared in our experiments.

\subsection{Point Cloud Compression}
Point cloud compression has been studied for years since a high-precision 3D point cloud model unavoidably takes huge storage space. According to \cite{cao20193d}, the common 3D point cloud compression method can be categorized into three groups: 1D traversal, 2D projection, and 3D correlation. For 1D traversal, \cite{gandoin2002progressive} is able to generate a connection list and build tree-based connectivity, which could help to predict adjacent points. Since this connectivity-driven approach represents points in 1D, geometric information is not well preserved. Thus, some methods project 3D onto 2D, e.g., projecting 3D point clouds to height fields \cite{ochotta2004compression} before using a shape-adaptive wavelet coder to encode. \cite{4378926} represents point clouds into several depth images according to different viewpoints, exploiting statistical dependencies from both temporal and inter-view reference pictures for prediction of both color and depth data. \cite{8571288} also converts point clouds into several separate video sequences, one for geometry and another for texture information, then using video codecs for compressing. Octree-based compression methods directly make use of 3D correlated information. \cite{huang2006octree} proposes to encode point clouds and their normals with arbitrary topology, they reorder the occupancy code to reduce the entropy. Binary tree and quadtree are combined in \cite{8486481} for effective compression.
Recently, inspired by image and video compression methods using deep learning, deep entropy model was introduced into point cloud compression. \cite{huang2020octsqueeze} utilizes octree to solve the sparsity problem of the LiDAR point cloud. It uses node’s tree structure information to reach a considerable compression ratio, while maintaining the same reconstruction accuracy compared with other algorithms by minimizing entropy distribution. Based on this, \cite{que2021voxelcontext} further exploits the voxel context in the octree-based framework by integrating neighboring nodes’ information which can be employed on both static and dynamic point clouds. However, no compression method also upsamples point clouds for higher density.

\section{Methodology}
The proposed CURL for the 3D point cloud representation mainly contains four parts: meshing, upsampling, encoding, and continuous reconstruction. Its pipeline is shown in Fig. \ref{fig:pipeline}.

\subsection{Meshing}
A single point cloud scan from a LiDAR sensor, particularly in outdoor environments, tends to be sparse with limited geometric resolution. This means it can cause inaccurate approximation for continuous reconstruction if directly encoding the sparse point cloud (See Section \ref{sec:PCencoding}). To tackle this challenge, CURL upsamples raw point cloud scans from a mesh representation before encoding.

Different from most popular point cloud mesh generation algorithms, which use 3D locations of points, we propose a polar-parametrized triangular meshing in a 2D space for three reasons: (1) this is much more efficient than meshing in the 3D space; (2) it is able to derive a convex hull to get a watertight mesh;  (3) triangular meshes linking points that are scanned with similar azimuth directions tend to be on an identical or nearby surface.

\begin{figure}[h]
\includegraphics[width=\linewidth]{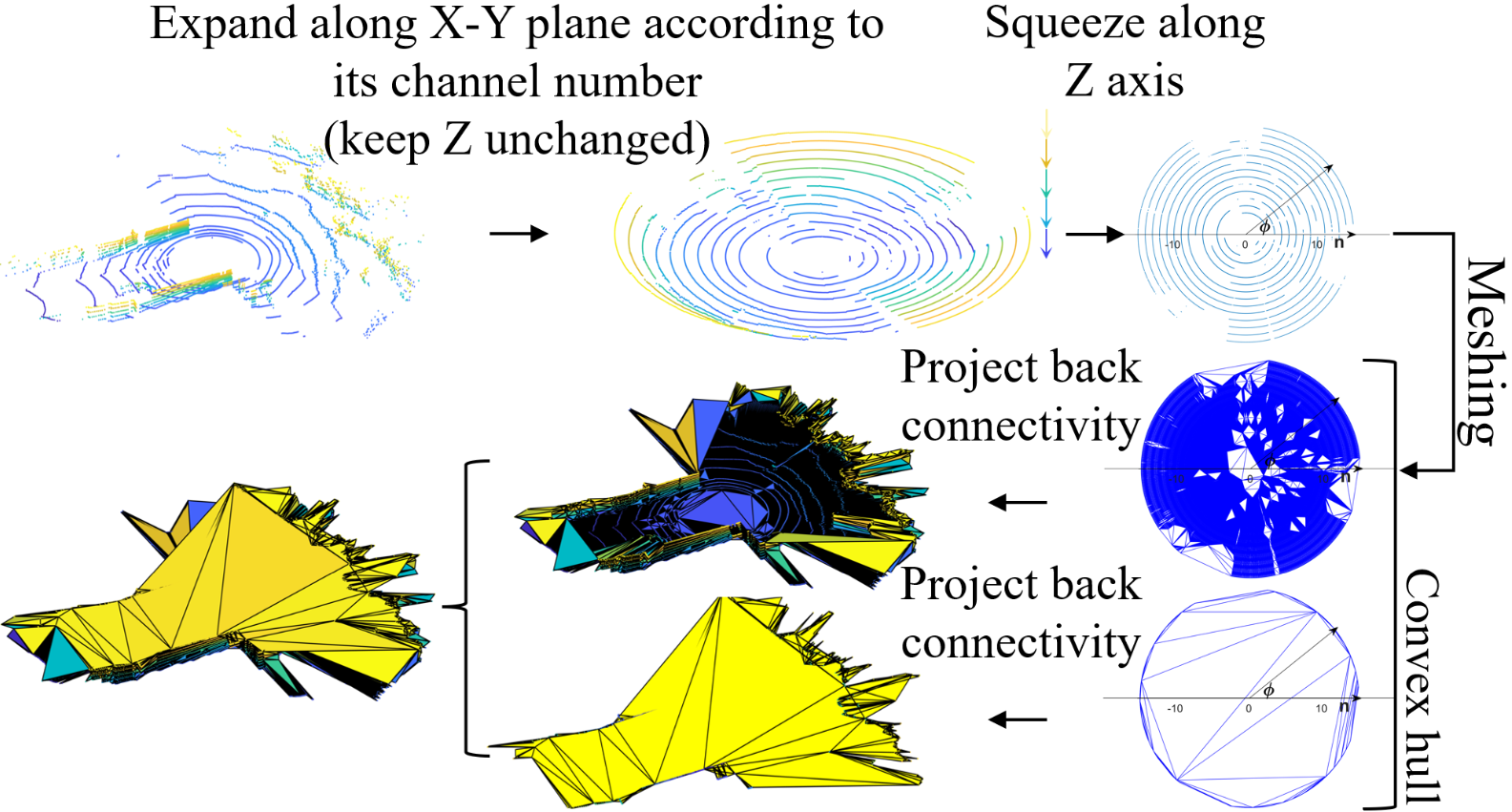}
\caption{Meshing pipeline.}
\label{fig:meshing}
\end{figure}

Fig. \ref{fig:meshing} shows an example of generating a mesh by using a 16-channel LiDAR point cloud. Points of a LiDAR scan are transferred into a two-dimensional polar coordinate $(n,\phi)$ using their channel order $n$ (as radius) and azimuth angle $\phi$. After performing this coordinate transformation, LiDAR points with similar shooting directions are gathered along the radial direction. Then, Delaunay triangulation \cite{baerentzen2012guide} method is utilized to produce a mesh for the 2D points. This process is very efficient since this is meshing in 2D, unlike point cloud meshing in 3D. Meanwhile, a convex closure is needed to make sure the mesh is watertight. After having these meshes of 2D vertices, their triangular connections can be simply mapped back into the corresponding 3D points to produce 3D triangular meshes of the point cloud efficiently.

\subsection{Upsampling}
Once having the meshes of the LiDAR scan, they are used to generate denser points via upsampling. Given the nature of upsampling and inaccuracy of meshing, some noisy points could be added during the upsampling process. Therefore, denoising is also introduced after sampling.

\subsubsection{Sampling}
For a physical 3D LiDAR sensor, its vertical and horizontal FoVs $\Theta_v$ and $\Theta_h$ and resolutions $\Delta_v$ and $\Delta_h$ are fixed by its design and configuration. Assume its number of vertical channel is $L$ and it has $N_h$ horizontal bins per scan. Without changing the sensor's configuration, it would be impossible to increase its FoVs and the number of channels and horizon bins. Therefore, sampling can mainly grow its point density by increasing its resolutions $\Delta_v$ and $\Delta_h$, which are key for long-range sensing (because the further away, the sparser point cloud would be). Based on this, a virtual LiDAR grid upsampling with higher resolutions $\Delta_v = \Theta_v / (L \cdot S_{row})$ and $\Delta_h = \Theta_h / (N_h \cdot S_{col})$, where $S_{row}$ and $S_{col}$ are increased sampling rates, is used to produce a denser point cloud from the generated mesh through ray tracking. The higher the resolutions are, the denser the upsampled point clouds are.



\subsubsection{Denoising with Masks} \label{sec:mask_application}
Upsampling can provide more geometric details based on the meshes, but meanwhile, noisy points are introduced during this process. A large number of noisy points can corrupt encoding, which leads to inaccurate and distorted reconstruction.
Therefore, we propose a denoising method based on masks that can be efficiently created to identify and remove noisy points.


\begin{figure}
    \centering
        \includegraphics[width=\linewidth]{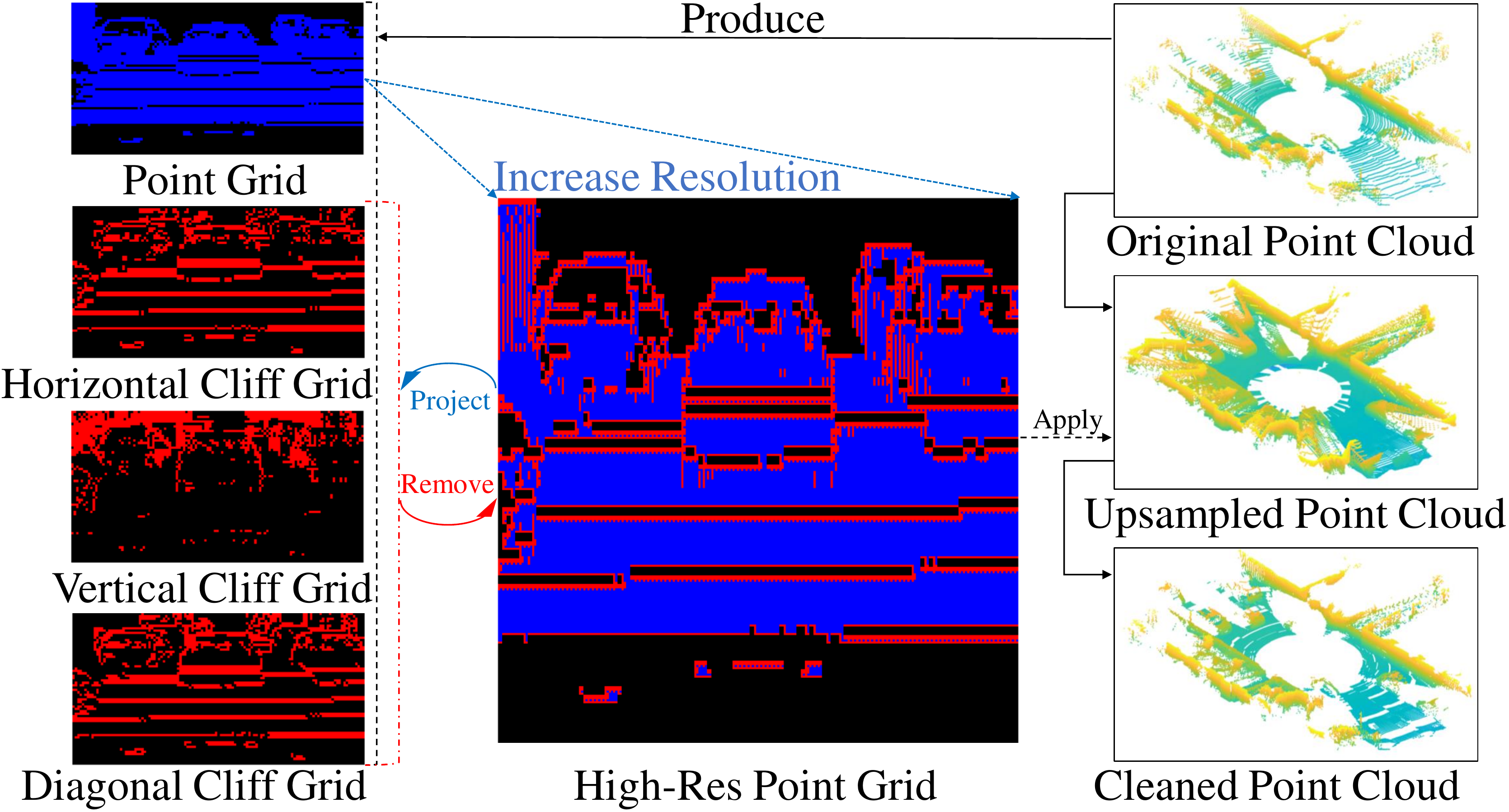}

     \caption{Mask generation and application. Four mask bitmaps are produced by using the depth image of a point cloud. 
     Cliff Grids are to detect wrong pixels (red pixels in the High-Res Point Grid). The High-Res Point Grid (blue pixels) can be used to filter noisy points generated from upsampling or continuous reconstruction results.}
\label{fig:sampling_and_mask}
\end{figure}

\paragraph{LiDAR Point Cloud Projection}\label{sec:PCProjection}
Since it is efficient to produce and apply masks on 2D images, the upsampled point cloud is projected as a depth image by transforming all its 3D points into a spherical coordinate system. The transformation is 
\begin{equation}
\label{sph_2_cart}
    \begin{split}
    	x &= r\cdot sin(\phi)cos(\theta) \\
        y &= r\cdot sin(\phi)sin(\theta) \\
        z &= r\cdot cos(\phi)
    \end{split}
\end{equation}
where $(x,y,z)$ is the Cartesian coordinate of a point and $(\theta, \phi, r)$ is its corresponding spherical coordinate.
We then map them to image space, as a depth image, with pixel height and width coordinates $(I_{h}, I_{w})$ calculated by:
\begin{equation}
\label{image_length}
    	I_{h} = (\theta + \Theta_v / 2) /\Delta_v,\quad
        I_{w}   = \phi/\Delta_h
\end{equation}
Therefore, each pixel location of the depth image corresponds to a spherical coordinate of the LiDAR, and its pixel intensity is the radial distance $r$. 

\paragraph{Mask Bitmaps}
Based on the depth image, the following bitmaps are produced.
\begin{itemize}
\item Point Grid: This bitmap is to identify whether there is a 3D point corresponding to a pixel.
Specifically, if a pixel has a valid depth, it is marked as $1$ to keep. Otherwise, it is $0$. Only the point grid yielded from the original point cloud is stored.
\item Cliff Grids: Sharp range changes happen for cliff points located at the foreground and background gaps. Hence, they are detected by examining the depth values of two adjacent pixels in horizontal, vertical, and diagonal directions. If they are very different along one direction, its neighbor pixel in that direction is marked as a cliff point to remove in the cliff bitmap. From Fig. \ref{fig:sampling_and_mask}, we can see that these mask grids support identifying wrong pixels of the High-Res Point Grid by projecting the point grid on the cliff grids to remove pixels out of the boundary.
\end{itemize}
By leveraging these mask bitmaps, we can then identify and remove most of the 3D outlier points in the upsampled point cloud. See the upsampled point cloud and the cleaned point cloud in Fig. \ref{fig:sampling_and_mask} as a comparison.

\subsection{Encoding}\label{sec:PCencoding}

\begin{figure}
\includegraphics[width=\linewidth]{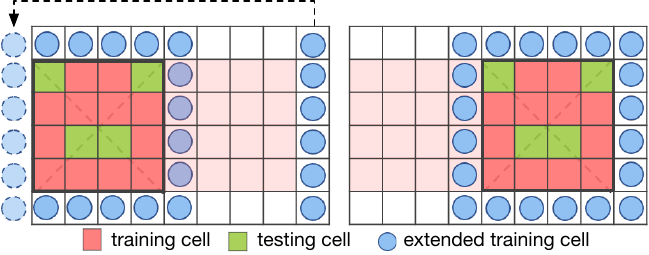}
\caption{Patching with training, testing and extended training cells. Left is one patch and right shows another patch. Patch size is $4\times4$ with $1$ extended training cell in this example.}
\label{fig:patch_moving}
\end{figure}

This encoding phase is to compute an adaptive spherical harmonics approximation of the upsampled point cloud. It consists of patching, spherical harmonics encoding, and adaptive refinement.

\subsubsection{Patching}
To improve encoding accuracy and computation efficiency, the depth image of the upsampled point cloud is divided into many small square patches for encoding. When performing encoding with a higher degree (See details in Section \ref{sec:SHEncoding}), the points are fitted more accurately. But it tends to overfit, causing more significant errors on unfitted points with less generalizability. Therefore, we introduce a strategy to validate how good a chosen encoding degree is to generalize in unknown areas by dividing the patch's cells into training and testing cells. As shown in Fig. \ref{fig:patch_moving}, the training cells (in red) are used to calculate encoding, while the testing cells (in green) are used to validate encoding errors. The testing cells are selected on the diagonal of the patch with each apart. Moreover, to build a correlation between neighbor patches for better consistency, each patch is padded with extended training cells (in blue), which are used together with the training cells for encoding. Note that the smaller the patch is, the faster the computation, the higher the accuracy, and the lower the compression ratio is.


\subsubsection{Spherical Harmonics Encoding}\label{sec:SHEncoding}

We are now ready to compute the encoding coefficients. In this work, the patches are encoded by using spherical harmonics (SPHARM), which approximately represents a 3D point cloud with a set of deformed spheres \cite{muller2006spherical}. Intuitively, it is the Fourier transform functions defined on the sphere. The spherical harmonics expansion function is defined as
\begin{equation}
\label{spherical_harmonics_expansion}
\begin{split}
	f(\theta,\phi) &= \sum_{l=0}^{\infty}\sum_{m = -l}^{l}c_{l}^{m}Y_{l}^{m}(\theta,\phi),\  (\theta \in[0,2\pi],\phi \in[0,\pi])\\
     l &= 0,1,2,3,\dots,\quad m =-l,-l+1,\dots,l-1,l
\end{split}
\end{equation}
i.e., a function $f(\theta,\phi)$ in the spherical coordinate can be represented by the linear combination of the spherical harmonics function $Y_{l}^{m}(\theta,\phi)$ of degree $l$ and order $m$. Since all points in a point cloud have been represented in the spherical coordinate in Section \ref{sec:PCProjection}, Eq. \eqref{spherical_harmonics_expansion} can be utilised to fit the spherical harmonics coefficient $c_l^m$ describing the points in a patch. The spherical harmonics function $Y_{l}^{m}(\theta,\phi)$ is defined as
\begin{equation}
\label{spherical_harmonics}
	Y_l^m(\theta,\phi) = (-1)^m\sqrt{\frac{(2l+1)}{4\pi}\frac{(l-m)!}{(l+m)!}}P_l^m(cos\theta)e^{im\phi}
\end{equation}
where $P_l^m(x)$ is associated Legendre polynomials \cite{muller2006spherical}. Therefore, if an upper limit $L_{max}$ of the degree $l$ is given, and an input function defined in spherical coordinate as $f(\theta,\phi)$ is described by a set of spherical samples $(\theta_i,\phi_i)$, then $f_i=f(\theta_i,\phi_i)$ where $1\leq i \leq n$, according to Eq. \eqref{spherical_harmonics_expansion} a linear system equation can be formulated as
\begin{equation}
\label{linear_system_matrix}
    \mathbf{YC}  = \mathbf{F}
    , \text{or}
	\begin{bmatrix}
    y_{1,1} & y_{1,2} & y_{1,3} & \cdots & y_{1,k}\\
    y_{2,1} & y_{2,2} & y_{2,3} & \cdots & y_{2,k}\\
    \vdots & \vdots & \vdots & & \vdots \\
    y_{n,1} & y_{n,2} & y_{n,3} & \cdots & y_{n,k}\\
	\end{bmatrix}
    \begin{bmatrix}
    c_{1} \\
    c_{2} \\
    c_{3} \\
    \vdots \\
    c_{k} \\
	\end{bmatrix}
     =
     \begin{bmatrix}
    f_{1} \\
    f_{2} \\
    \vdots \\
    f_{n} \\
	\end{bmatrix}
\end{equation}
, where $y_{i,j} = Y_l^m(\theta_i,\phi_i)$, $j = l^2+l+m+1$ and $k = (L_{max}+1)^2$. Note that we use an indexing scheme that assigns a unique index $j$ to every pair $(l,m)$. The least square method can then be employed to optimally calculate the spherical harmonics coefficients. 
Since $f(\theta,\phi)$ is defined as a continuous function, the point cloud encoding is also continuous.
This means once the spherical harmonics coefficients are derived, they can be saved or transmitted to reconstruct the original point cloud continuously. Since these coefficients are only a few scale values, they are much more efficient and compact in terms of memory and storage sizes compared with the original point cloud that contains thousands of 3D points.

\subsubsection{Adaptive Refinement}
Because the geometry of environments can be simple or complex, the optimal SPHARM degree for each patch may vary significantly. If the degree is too high, the reconstruction using the encoded SPHARM coefficients would be inaccurate, and it takes more space to store. On the other hand, if the degree is too low, even the training cells may not be well fitted. Therefore, an algorithm is designed to select a suitable degree to balance these two situations adaptively.

\begin{figure}[b]
    \centering
    \includegraphics[width=0.6\linewidth]{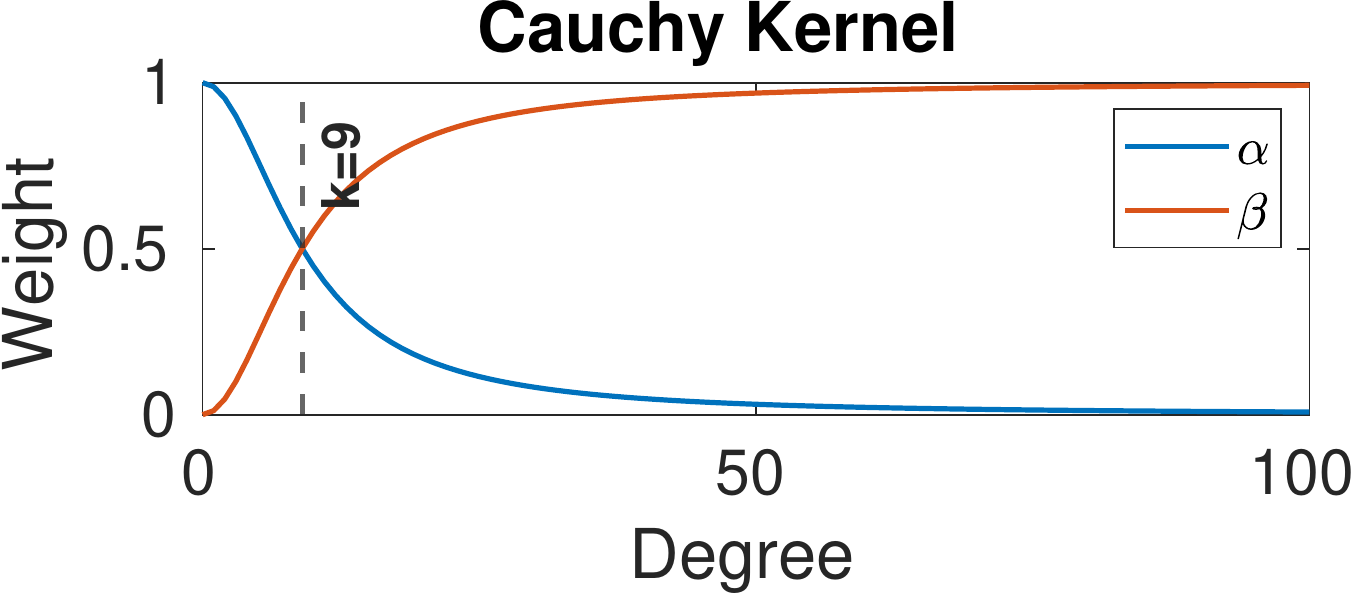}
    \caption{$\alpha$ and $\beta$ change with respect to SPHARM degree, and equal when $k=9$.}
    \label{fig:cauchy_kernel}
\end{figure}

Since after performing spherical harmonics encoding using the training and extended training cells, we can utilize the derived SPHARM coefficients to recover the points in the training cells and testing cells. Based on this, optimization is built to optimize the degree, iteratively minimizing the errors between the recovered points and their real values.
Denote $\alpha$ and $\beta$ as the weights for the training and testing errors, where

\begin{equation}
    \label{eq:cauchy kernel}
        \alpha = W, \quad
        \beta = 1-W, \quad
        W = \frac{1}{1+(\frac{l}{k})^2}
\end{equation}

\begin{equation}
    \label{eq:error equation}
    E_{t} = \alpha E_{a}+\beta E_{b}
\end{equation}
$E_a$ is the training error, $E_b$ is the testing error, and $E_t$ is the total error. Hence, when $l = k$, $\alpha = \beta$ which means the error weights of the training and testing cells equal. According to Fig. \ref{fig:cauchy_kernel}, we empirically select $k = 9$, i.e., when the degree smaller than $9$, we value the errors of training cell more. Once the average total error for each patch in Eq. \eqref{eq:error equation} is smaller than a certain threshold or starting to increase, the iteration is finished.

\subsection{Continuous Reconstruction \label{sec:continuous_recons}}
Now all the SPHARM coefficients can be used for continuous reconstruction.
To perform dense reconstruction, a fine grid $(\theta_{r}, \phi_{r})$ is generated to retrieve 3D points. 
Eq. \eqref{spherical_harmonics} is employed to calculate $\mathbf{Y}$ in Eq. \eqref{linear_system_matrix}. Since the SPHARM coefficients have been derived before, $\mathbf{F}$ which is the set of radius distances corresponding to the points in the grid $(\theta_{r}, \phi_{r})$ can be computed efficiently. Finally, the mask bitmaps can be applied to remove noisy points that are newly reconstructed if preferred. Depending on the desired density of the point cloud to reconstruct, the resolution of the grid can be adjusted accordingly. Since the resolution can be changed continuously, the density of the reconstruction is also continuous.
Fig. \ref{fig:comparsion} shows two reconstruction results with different densities.



\begin{table}[]
    \centering
\begin{tabular}{|c|c|c|c|}
\hline
\textbf{Name}                             & \textbf{Variable}          & \textbf{Indoor}             & \textbf{Outdoor}            \\ \hline
\textbf{Scan Channels}                    & $L$                        & $64$                        & $64$                        \\ \hline
\multirow{2}{*}{\textbf{Sampling Rate}}   & $S_{row}$                  & $2$                         & $2$                         \\ \cline{2-4}
                                          & $S_{col}$                  & $2$                         & $2$                         \\ \hline
\multirow{3}{*}{\textbf{Cliff Threshold}} & Horizontal & 0.1 m     &  2.0 m       \\\cline{2-4}
                                          & Vertical                   & 0.1 m                       & 0.2 m                      \\ \cline{2-4}
                                          & Diagonal                   & 0.1414 m  & 2.0 m \\ \hline
\multirow{3}{*}{\textbf{Patch Size}}      & $P_r$                      & $4$ pixels                         & $4$ pixels                        \\ \cline{2-4}
                                          & $P_{row}$                  & $P_r\times S_{row}$         & $P_r\times S_{row}$         \\ \cline{2-4}
                                          & $P_{col}$                  & $P_r\times S_{col}$         & $P_r\times S_{col}$         \\ \hline
\multirow{2}{*}{\textbf{Reconstruction}}  & $R_{row}$                  & $1$                         & $1$                         \\ \cline{2-4}
                                          & $R_{col}$                  & $1$                         & $1$                         \\ \hline
\end{tabular}
    \caption{Default CURL parameters.}
    \label{tab:parameters}
\end{table}

\section{Experiments}
In this section, we evaluate the proposed CURL representation on four public datasets: Newer College \cite{ramezani2020newer}, KITTI \cite{Geiger2012CVPR}, Indoor LiDAR-RGBD SCAN \cite{Park2017} and SuperRso \cite{shan2020simulation} datasets. Their scenarios include college gardens, street scenes, and indoor environments with different LiDAR sensors and degrees of structural complexity.
The LiDAR sensor used in the Newer College Dataset is an Ouster OS-1 (Gen 1) 64, combined with a Leica BLK360, which produces dense point cloud as ground truth reconstruction. Therefore, this dataset is suitable for evaluating the reconstruction accuracy. KITTI dataset uses a Velodyne HDL-64E without ground truth reconstruction. Thus, it is mainly used for assessing $1:1$ reconstruction. For the indoor dataset, a survey-grade FARO Focus 3D X330 scanner is used to scan $360^{\circ}$ vertical and horizontal fields of view within $3mm$ ranging accuracy. We also use it to evaluate the reconstruction accuracy quantitatively. Given that the changes on consecutive scans tend to be unnoticeable, LiDAR scans for encoding and reconstruction (not ground truth) are selected $5$ meters apart. 
The quantitative precision evaluation is conducted by comparing distances between reconstructed points and their nearest neighbors in the ground truth like \cite{huang2020octsqueeze}. For compression evaluation, we borrow the definition of compression percentage in the Point Cloud Library library (PCL) \cite{5980567}, which is bytes per compressed point divided by bytes per uncompressed point. Similarly, we define the compression percentage as bytes per compressed point cloud divided by bytes per uncompressed point cloud, where in Eq. \eqref{eq:compression rate} $C$ is the size of a compressed point cloud and $U$ is the size of an uncompressed point cloud.
\begin{equation}
    \label{eq:compression rate}
    CP = \frac{C}{U}\times 100\%
\end{equation}


\begin{figure}
    \centering
    \begin{subfigure}[b]{0.48\linewidth}
        \centering
        \includegraphics[width=\linewidth]{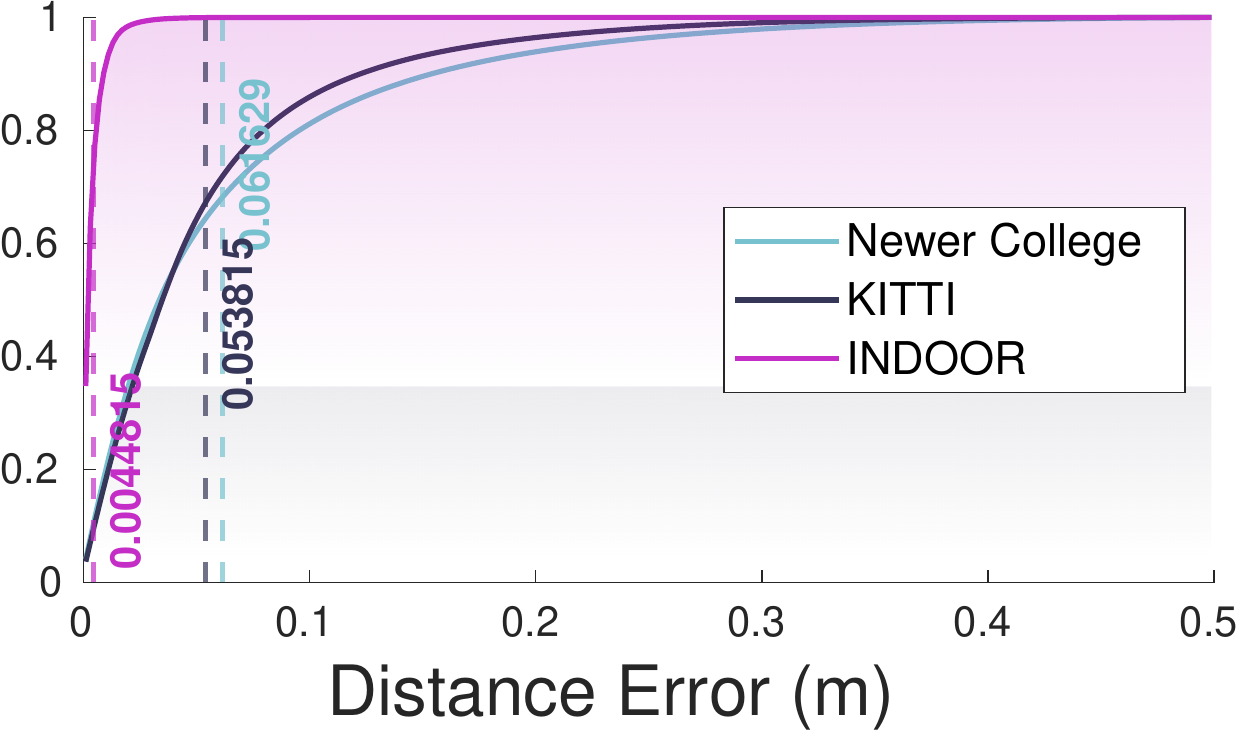}
        \caption{Reconstruction error}
        \label{fig:Error threshold avg a}
    \end{subfigure}
    \hfill
    \begin{subfigure}[b]{0.48\linewidth}
        \centering
        \includegraphics[width=\linewidth]{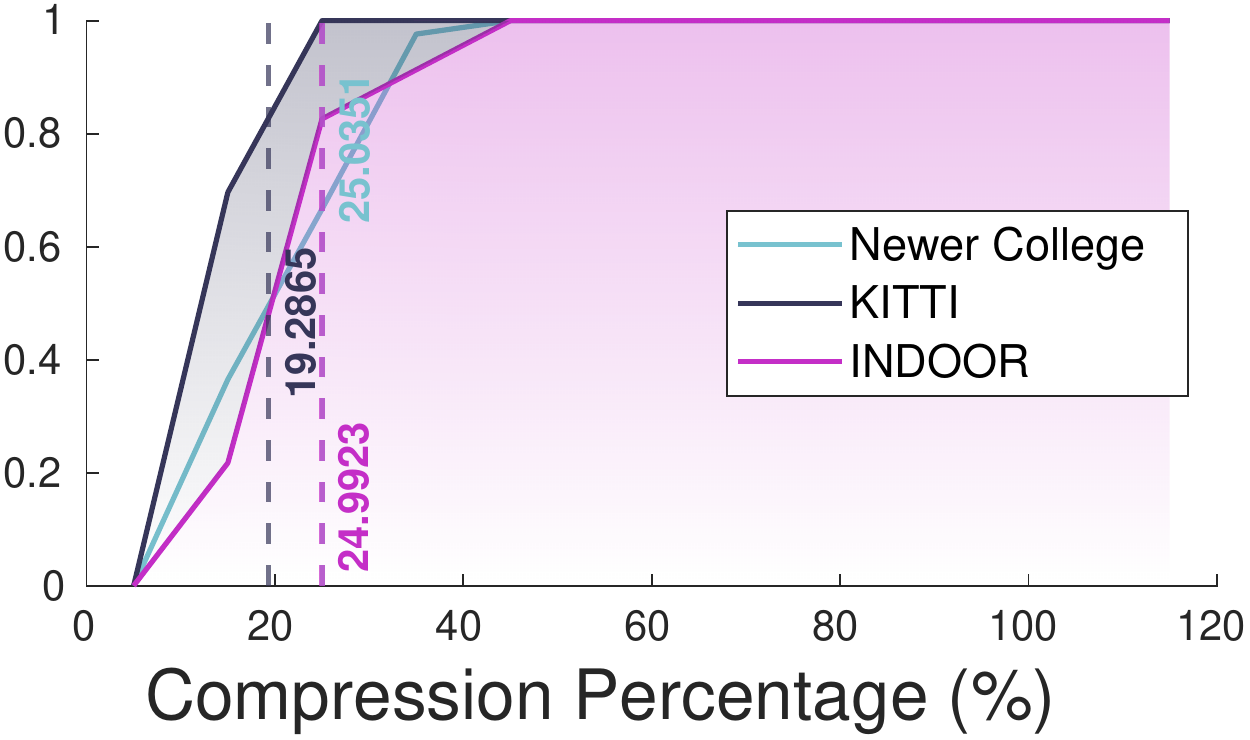}
        \caption{Compression Percentage}
        \label{fig:Error threshold avg b}
    \end{subfigure}
       \caption{Cumulative distribution of CURL's reconstruction errors and compression percentage for 1:1 reconstruction.}
       \label{fig:all datasets}
\end{figure}

\begin{figure}
   \centering


   \begin{subfigure}[b]{0.24\columnwidth}
   \centering
   \includegraphics[height=2cm,width=\textwidth]{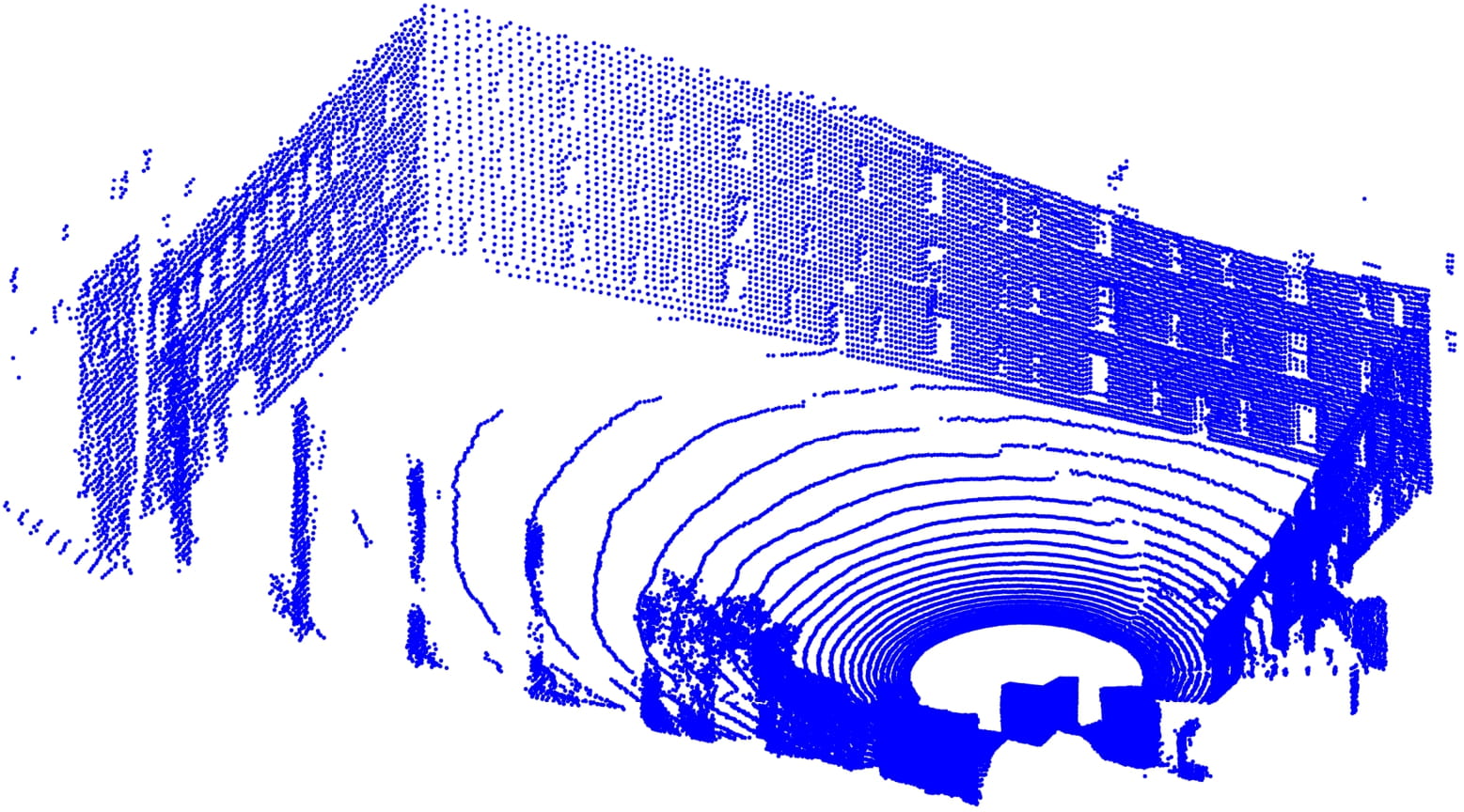}
   \end{subfigure}
   \begin{subfigure}[b]{0.24\columnwidth}
   \centering
   \includegraphics[height=2cm,width=\textwidth]{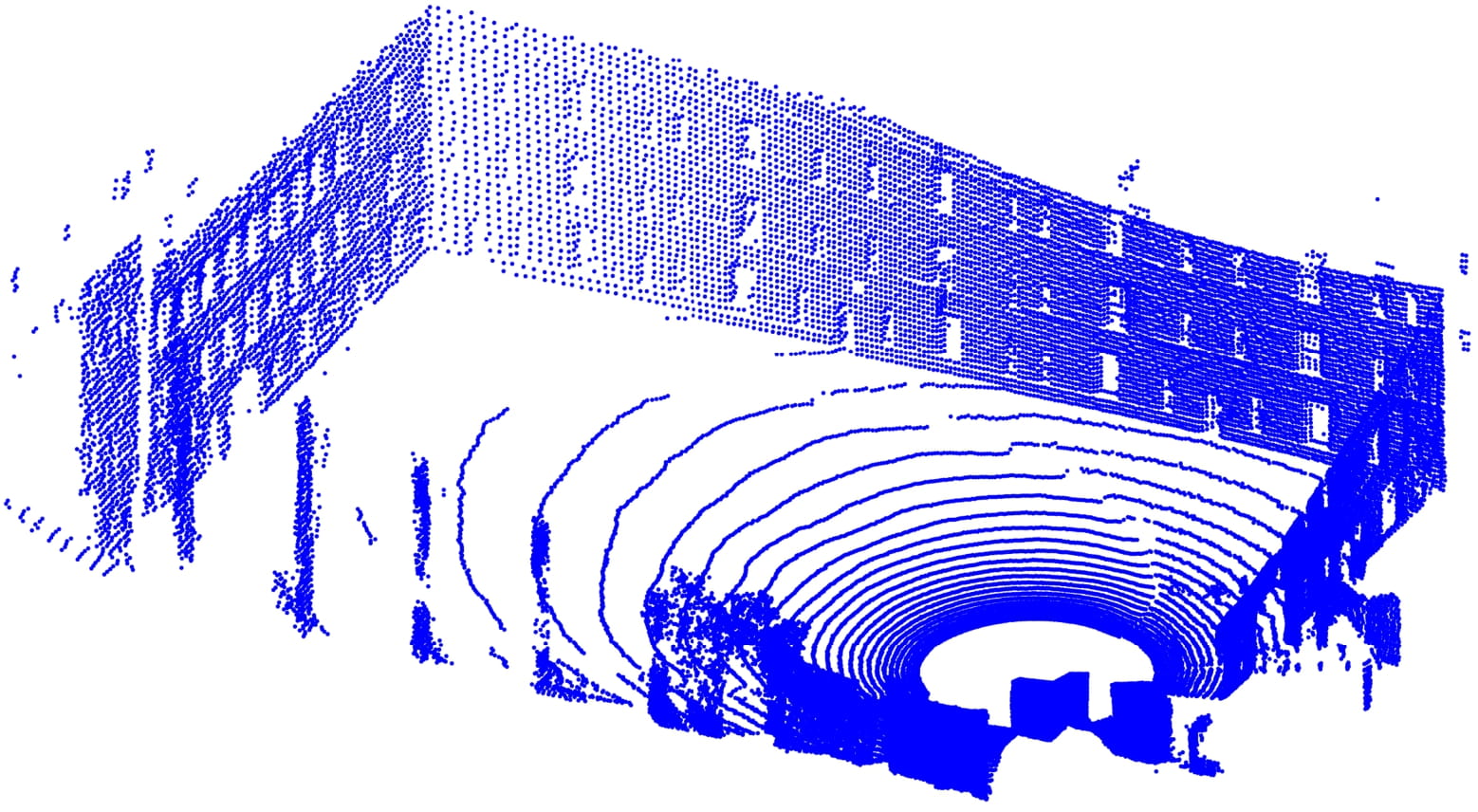}
   \end{subfigure}
   \begin{subfigure}[b]{0.24\columnwidth}
   \centering
   \includegraphics[height=2cm,width=\textwidth]{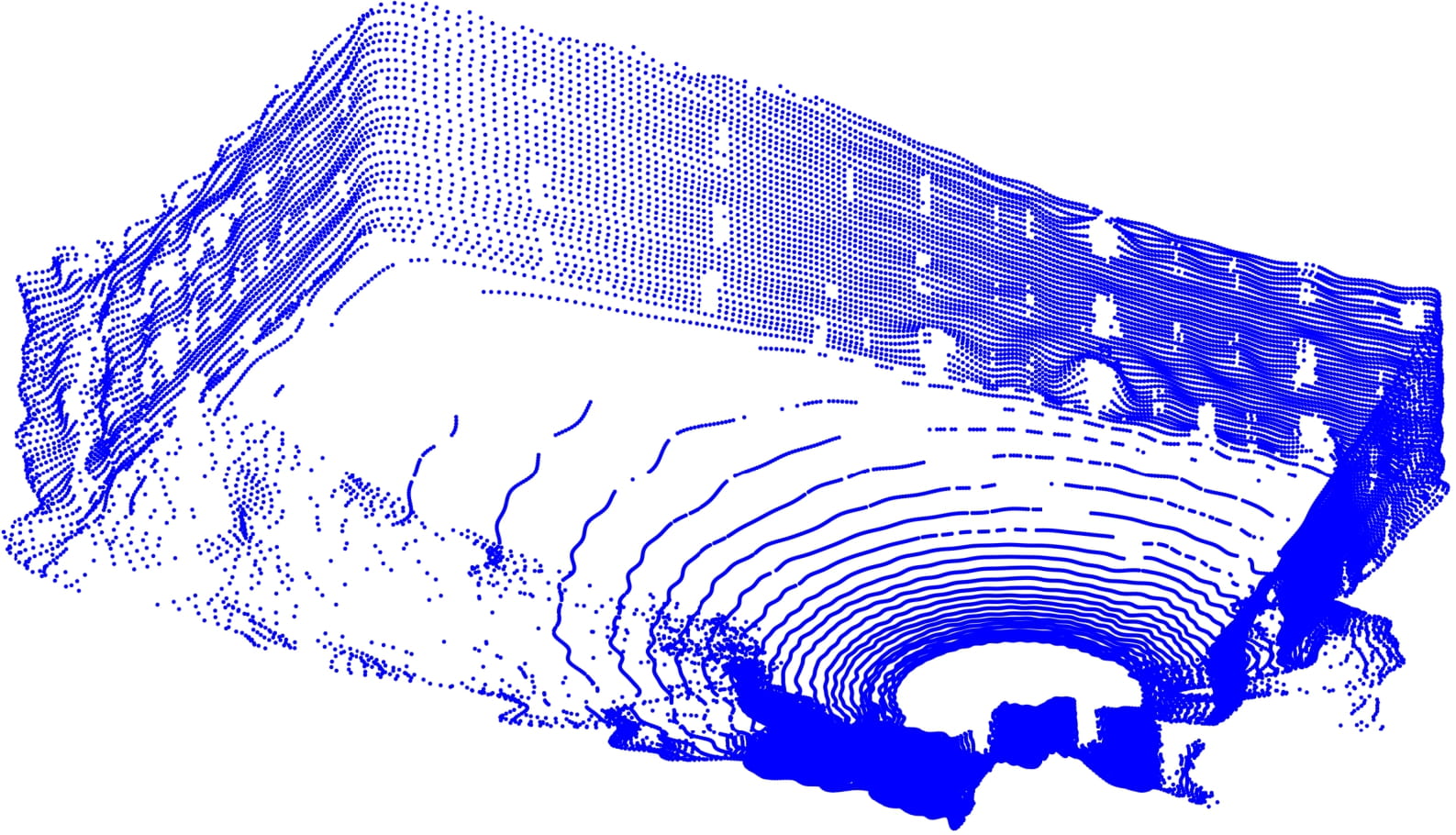}
   \end{subfigure}
   \begin{subfigure}[b]{0.24\columnwidth}
   \centering
   \includegraphics[height=2cm,width=\textwidth]{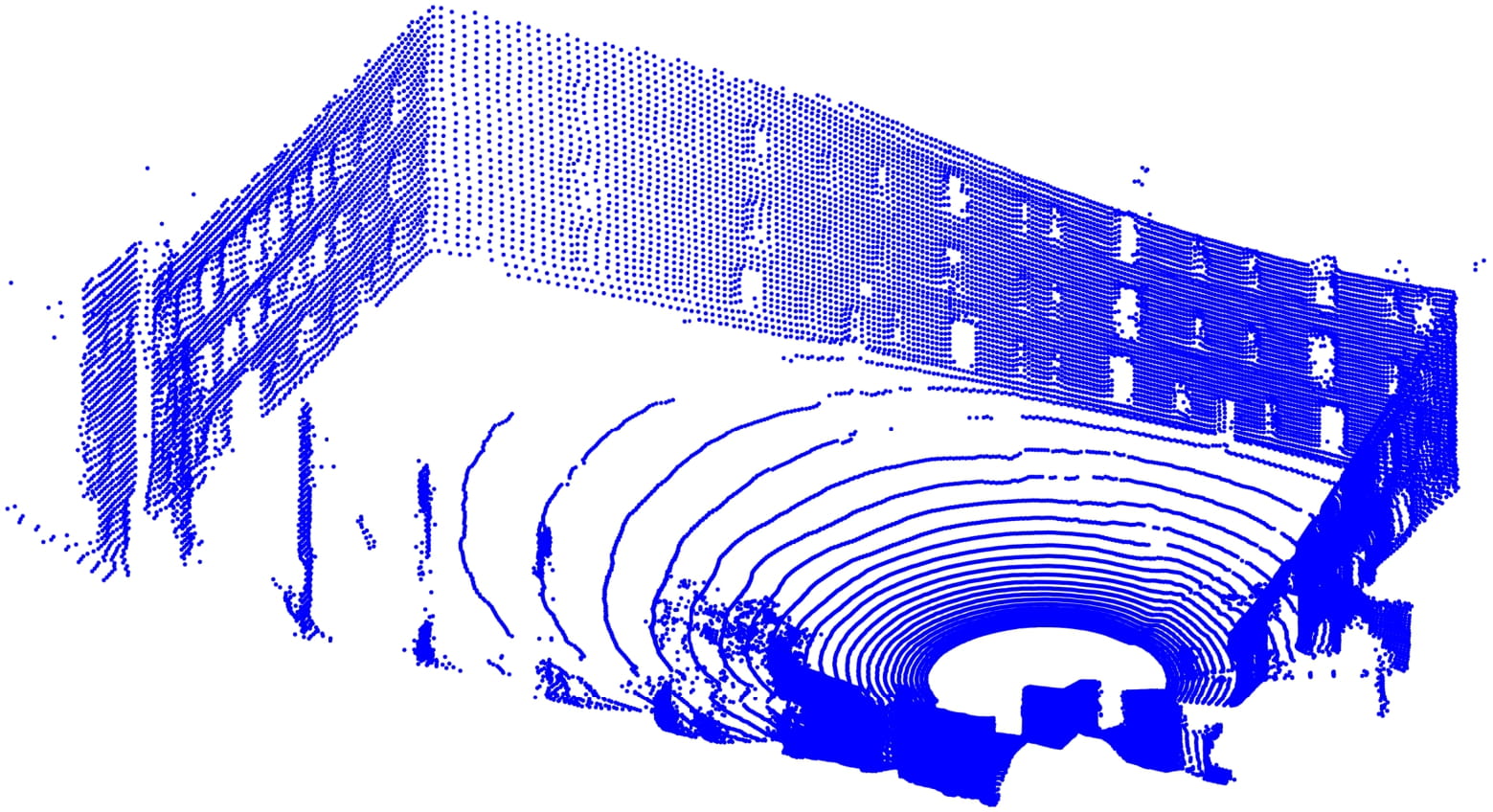}
   \end{subfigure}




   \begin{subfigure}[b]{0.24\columnwidth}
   \centering
   \includegraphics[height=2cm,width=\textwidth]{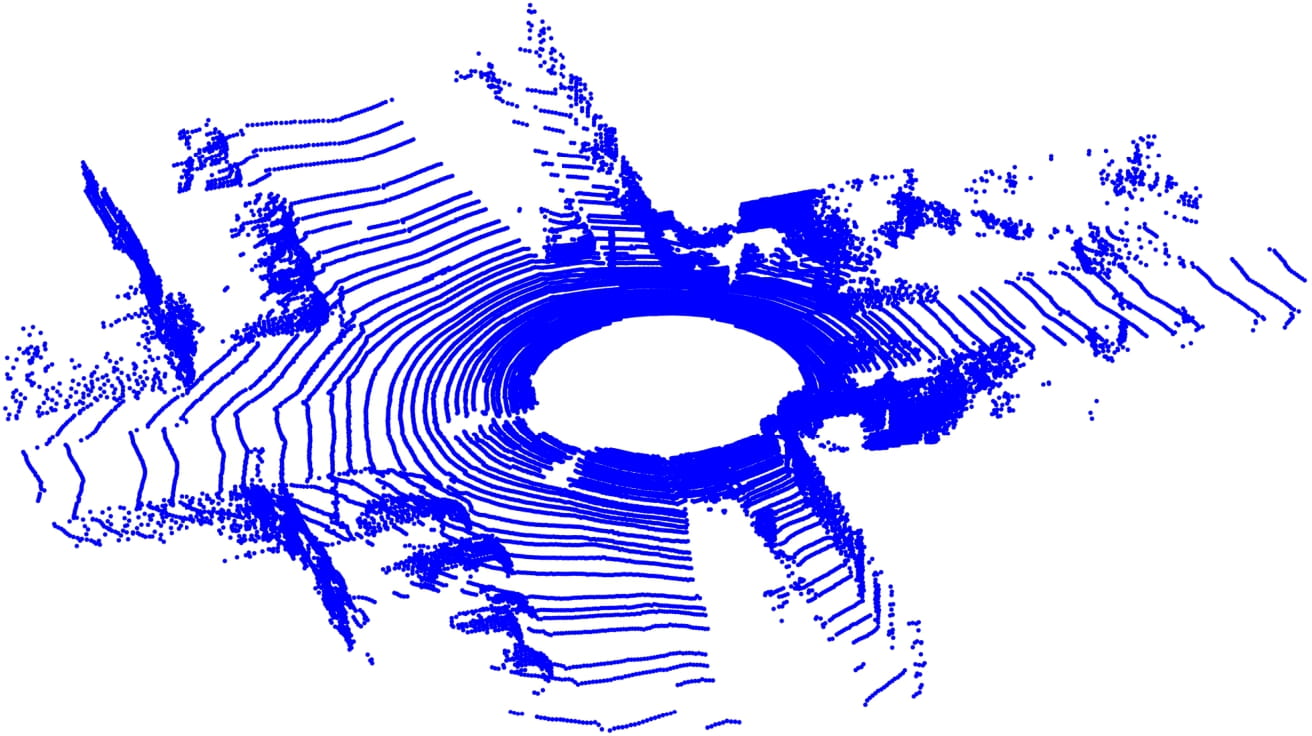}
   \end{subfigure}
   \begin{subfigure}[b]{0.24\columnwidth}
   \centering
   \includegraphics[height=2cm,width=\textwidth]{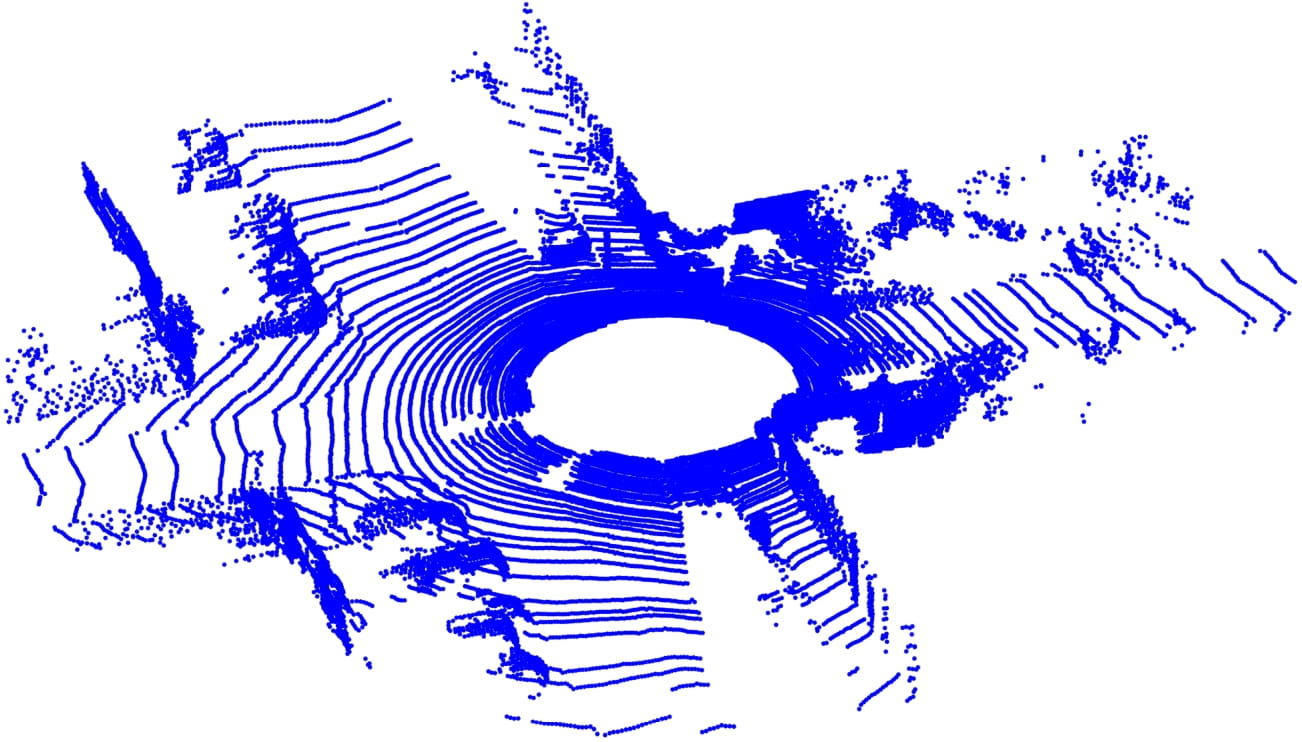}
   \end{subfigure}
   \begin{subfigure}[b]{0.24\columnwidth}
   \centering
   \includegraphics[height=2cm,width=\textwidth]{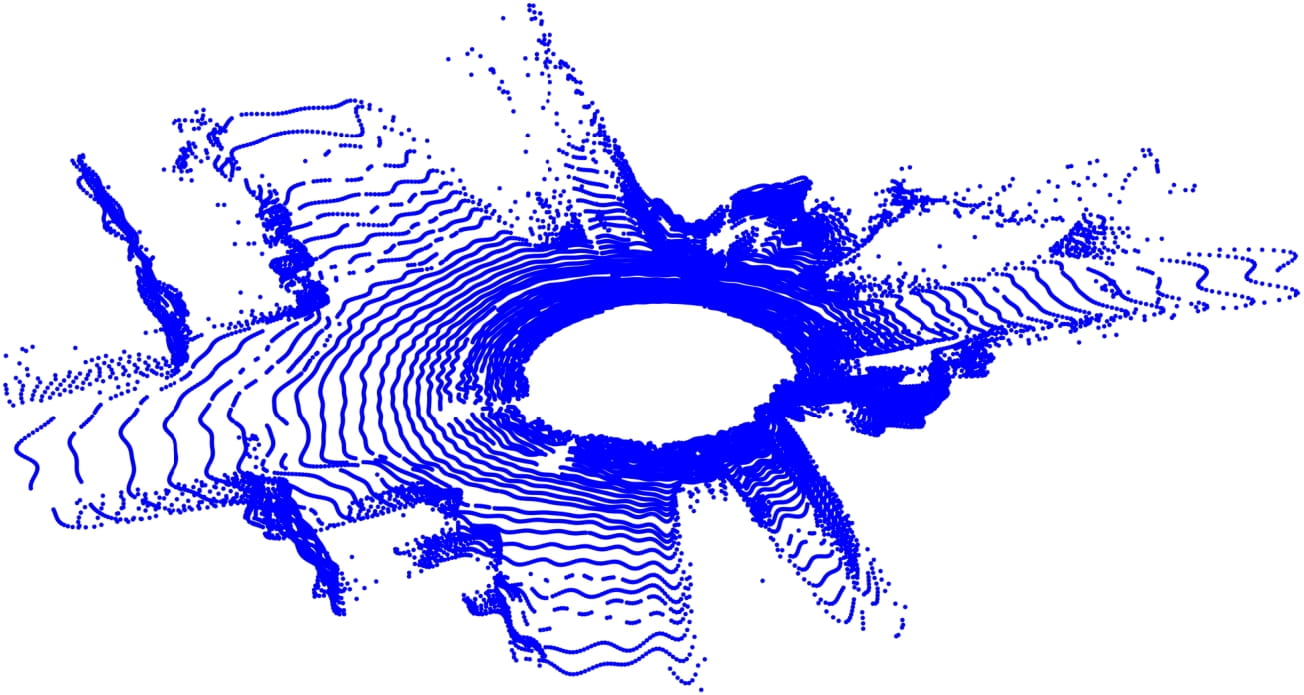}
   \end{subfigure}
   \begin{subfigure}[b]{0.24\columnwidth}
   \centering
   \includegraphics[height=2cm,width=\textwidth]{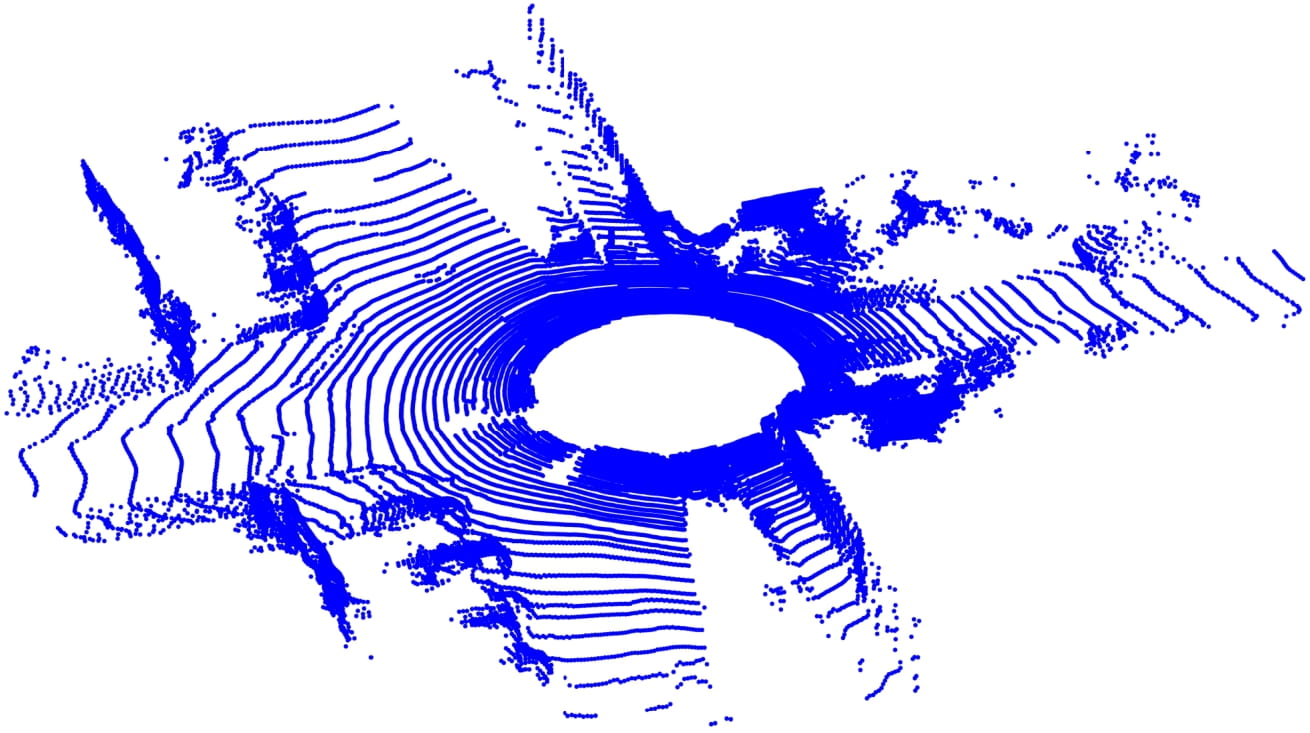}
   \end{subfigure}


   \begin{subfigure}[b]{0.24\columnwidth}
   \centering
   \includegraphics[height=2cm,width=\textwidth]{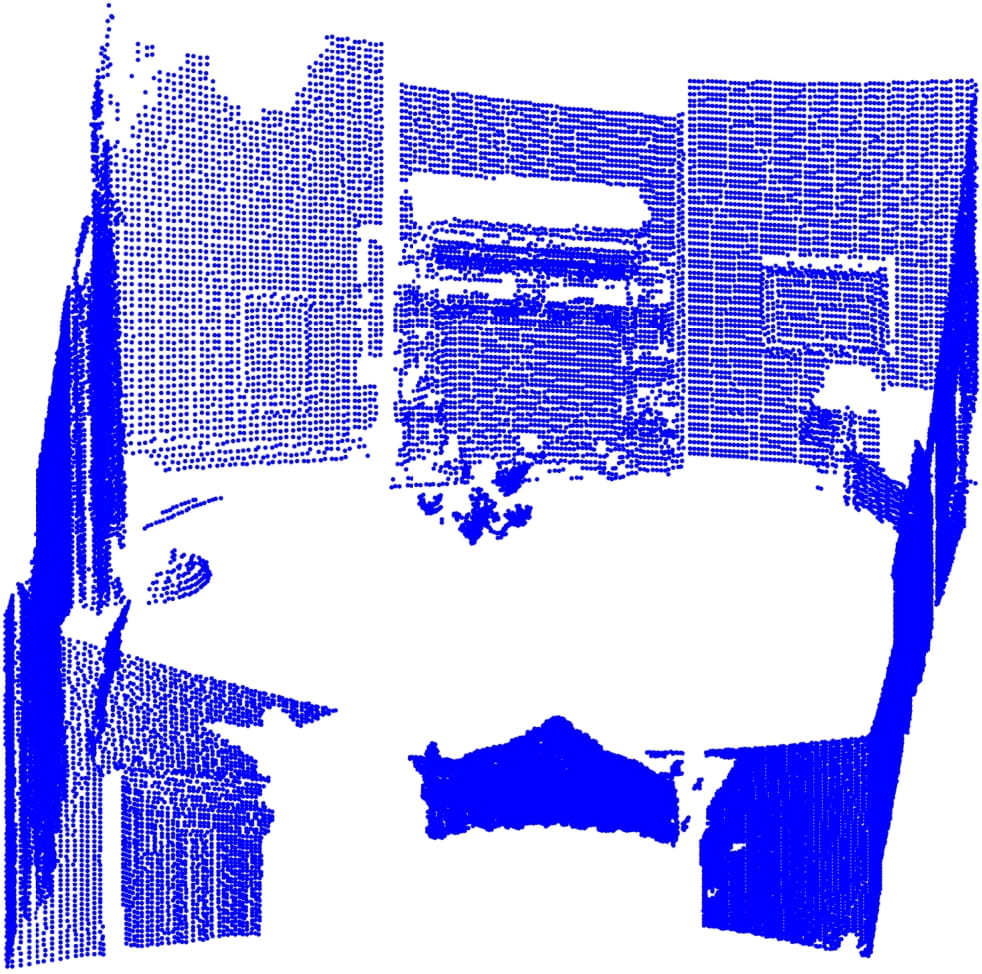}
   \caption*{\textbf{GT}}
   \end{subfigure}
   \begin{subfigure}[b]{0.24\columnwidth}
   \centering
   \includegraphics[height=2cm,width=\textwidth]{figures/tb1_INDOOR_0_4_PCL_LOW_ONLINE_c-1.jpg}
   \caption*{\textbf{PCL}}
   \end{subfigure}
   \begin{subfigure}[b]{0.24\columnwidth}
   \centering
   \includegraphics[height=2cm,width=\textwidth]{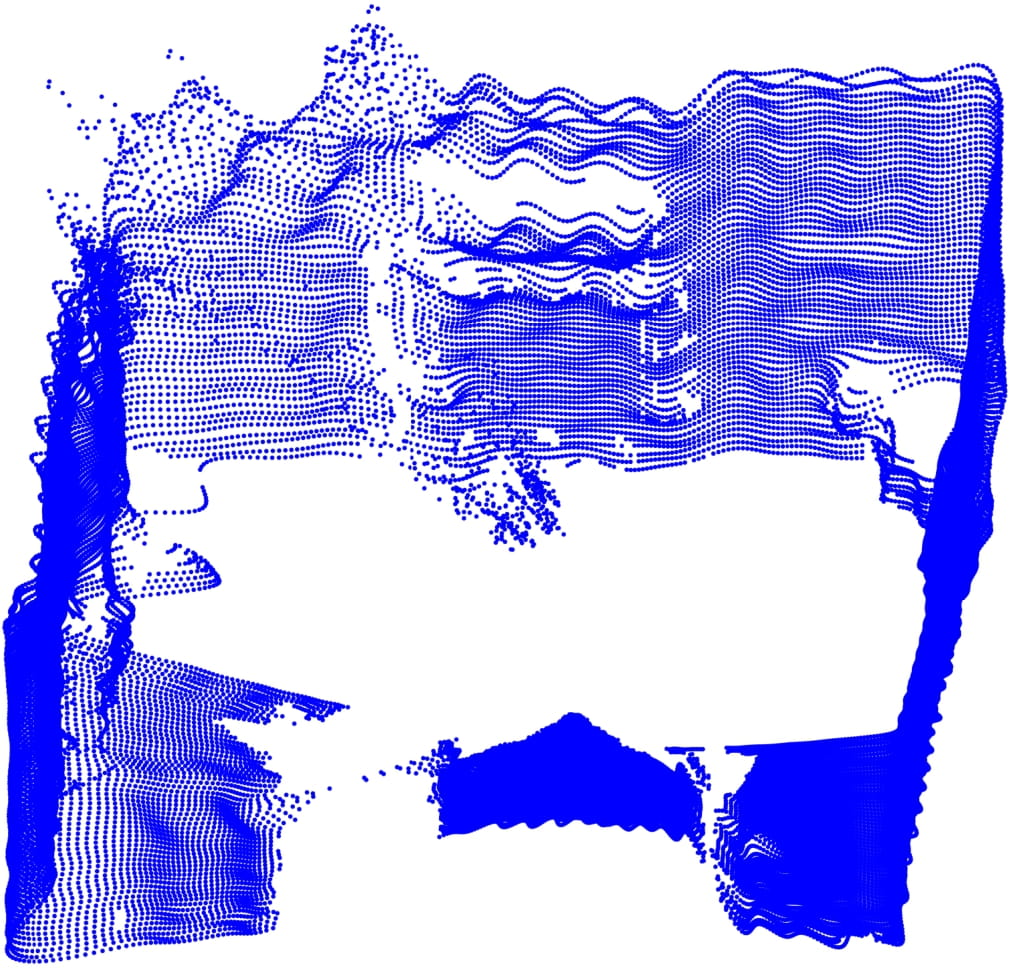}
   \caption*{\textbf{Baseline}}
   \end{subfigure}
   \begin{subfigure}[b]{0.24\columnwidth}
   \centering
   \includegraphics[height=2cm,width=\textwidth]{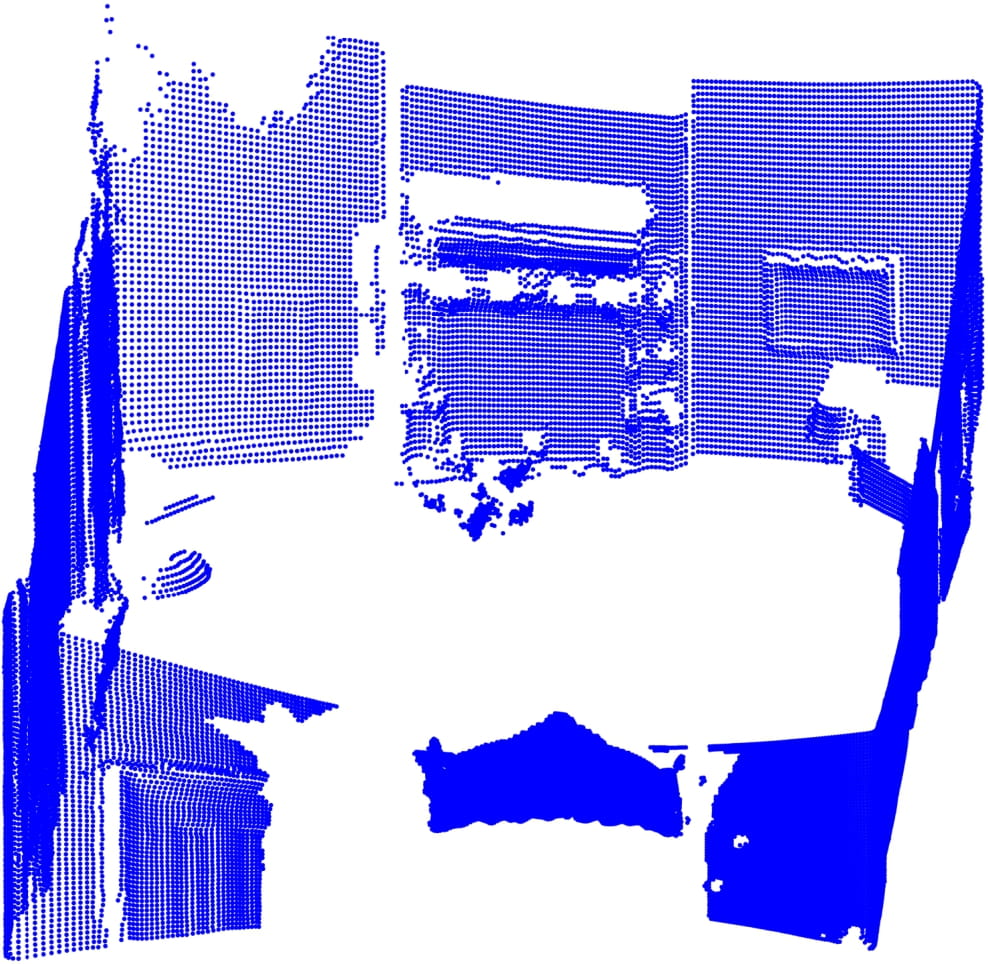}
   \caption*{$\mathbf{Our (Pr=2)}$}
   \end{subfigure}

   \caption{Examples of 1:1 reconstruction. Top to bottom: Newer College, KITTI and Indoor datasets. PCL is online mode with low resolution.
   }
   \label{fig:1:1 reconstruction for three datasets}
\end{figure}

\subsection{Parameters of CURL}
TABLE \ref{tab:parameters} shows CURL's parameters with their default values.
The default scans channel $L$ is set $64$ unless otherwise stated.
$P_r$ is the patch size that contains $P_r\times P_r$ pixels in an original depth image.
Since the depth image after upsampling contains more pixels, $P_{row}$ and $P_{col}$ represent the patch size of the upsampled depth image.
$R_{row}$ and $R_{col}$ determines desired multipliers on increasing row and column resolutions for continuous reconstruction. Their default values are $1$ as set for 1:1 reconstruction.


\subsection{Competing Methods}
Given that the point cloud compression and density increase for large-scale LiDAR scans are relatively under-explored for robotic applications, the following publicly available methods are selected for comparison.
For the 1:1 reconstruction experiment, we compare CURL with the point cloud compression method in the PCL. It is an octree-based method with online and offline modes. Since it only does point cloud compression without increasing density, it is not compared to upsampling and continuous reconstruction experiments.
Instead, LiDAR super-resolution (SuperRso) \cite{shan2020simulation}, a deep learning-based method that increases the density of point cloud from a LiDAR sensor, is benchmarked for upsampling. We use its official pre-trained weights that upsample a 16-channel scan to a 64-channel. Given that SuperRso does not report results of denser point clouds over 64 channels, it is only used to evaluate upsampling from a 16-channel to a 64-channel.
To evaluate continuous reconstruction, we employ a basic $64$-degree SPHARM method as a baseline without patching and adaptive refinement since it is the traditional SPHARM shape modeling method used in \cite{shen2006large}, \cite{shen2006spherical}, and \cite{brechbuhler1995parametrization}.


\subsection{Evaluation on 1:1 Reconstruction} \label{section:1:1 reconstruction}
The 1:1 reconstruction task compresses a point cloud and later recovers it without increasing its density. Therefore, the ground truth point cloud is the original point cloud captured from a LiDAR sensor. The 1:1 reconstruction experiments aim to understand the performance of a point cloud compression and recovery.

\begin{table}[]
\centering
\resizebox{\columnwidth}{!}{

\begin{tabular}{|cc|cccccccc|}
\hline
\multicolumn{2}{|c|}{\multirow{4}{*}{Dataset}}              & \multicolumn{8}{c|}{Method}                                                                                                                                                                                                                                                                      \\ \cline{3-10}
\multicolumn{2}{|c|}{}                                      & \multicolumn{1}{c|}{\multirow{3}{*}{Baseline}} & \multicolumn{6}{c|}{PCL}                                                                                                                                                                                                & \multirow{3}{*}{\shortstack{CURL\\(Ours)}} \\ \cline{4-9}
\multicolumn{2}{|c|}{}                                      & \multicolumn{1}{c|}{}                          & \multicolumn{3}{c|}{Online}                                                                              & \multicolumn{3}{c|}{Offline}                                                                                 &                       \\ \cline{4-9}
\multicolumn{2}{|c|}{}                                      & \multicolumn{1}{c|}{}                          & \multicolumn{1}{c|}{low}     & \multicolumn{1}{c|}{mid}     & \multicolumn{1}{c|}{high}                  & \multicolumn{1}{c|}{low}       & \multicolumn{1}{c|}{mid}       & \multicolumn{1}{c|}{high}                  &                       \\ \hline
\multicolumn{1}{|c|}{\multirow{3}{*}{\shortstack{Newer\\College}}} & mean & \multicolumn{1}{c|}{0.165}                     & \multicolumn{1}{c|}{0.005}   & \multicolumn{1}{c|}{0.005}   & \multicolumn{1}{c|}{0.0001}                & \multicolumn{1}{c|}{$0.005$}   & \multicolumn{1}{c|}{$0.002$}   & \multicolumn{1}{c|}{$4.8\mathrm{e}{-05}$}  & $0.062$               \\ \cline{2-10}
\multicolumn{1}{|c|}{}                               & std  & \multicolumn{1}{c|}{0.380}                     & \multicolumn{1}{c|}{0.001}   & \multicolumn{1}{c|}{0.001}   & \multicolumn{1}{c|}{$2.7\mathrm{e}{-05}$}  & \multicolumn{1}{c|}{$0.001$}   & \multicolumn{1}{c|}{$0.001$}   & \multicolumn{1}{c|}{$1.4\mathrm{e}{-05}$}  & $ 0.071$              \\ \cline{2-10}
\multicolumn{1}{|c|}{}                               & CP   & \multicolumn{1}{c|}{8.11\%}                    & \multicolumn{1}{c|}{16.76\%} & \multicolumn{1}{c|}{20.16\%} & \multicolumn{1}{c|}{$37.78\%$}             & \multicolumn{1}{c|}{$16.76\%$} & \multicolumn{1}{c|}{$20.10\%$} & \multicolumn{1}{c|}{$38.46\%$}             & $25.03\%$             \\ \hline
\multicolumn{1}{|c|}{\multirow{3}{*}{KITTI}}         & mean & \multicolumn{1}{c|}{0.076}                     & \multicolumn{1}{c|}{0.005}   & \multicolumn{1}{c|}{0.005}   & \multicolumn{1}{c|}{$0.0002$}              & \multicolumn{1}{c|}{$0.005$}   & \multicolumn{1}{c|}{$0.002$}   & \multicolumn{1}{c|}{$4.8\mathrm{e}{-05}$}  & $0.053$               \\ \cline{2-10}
\multicolumn{1}{|c|}{}                               & std  & \multicolumn{1}{c|}{0.105}                     & \multicolumn{1}{c|}{0.007}   & \multicolumn{1}{c|}{0.007}   & \multicolumn{1}{c|}{$0.007$}               & \multicolumn{1}{c|}{$0.001$}   & \multicolumn{1}{c|}{$0.0007$}  & \multicolumn{1}{c|}{$1.4\mathrm{e}{-05}$}  & $0.060$               \\ \cline{2-10}
\multicolumn{1}{|c|}{}                               & CP   & \multicolumn{1}{c|}{9.11\%}                    & \multicolumn{1}{c|}{14.90\%}  & \multicolumn{1}{c|}{18.35\%} & \multicolumn{1}{c|}{$36.14\%$}               & \multicolumn{1}{c|}{$14.90\%$} & \multicolumn{1}{c|}{$18.30\%$} & \multicolumn{1}{c|}{$36.78\%$}             & $19.29\%$              \\ \hline
\multicolumn{1}{|c|}{\multirow{3}{*}{INDOOR}}        & mean & \multicolumn{1}{c|}{0.0234}                     & \multicolumn{1}{c|}{0.005}   & \multicolumn{1}{c|}{0.005}   & \multicolumn{1}{c|}{$9.6\mathrm{e}{-05}$}  & \multicolumn{1}{c|}{$0.005$}   & \multicolumn{1}{c|}{$0.002$}   & \multicolumn{1}{c|}{$4.8\mathrm{e}{-05}$}  & $0.004$               \\ \cline{2-10}
\multicolumn{1}{|c|}{}                               & std  & \multicolumn{1}{c|}{0.0622}                    & \multicolumn{1}{c|}{0.0014}  & \multicolumn{1}{c|}{0.001}   & \multicolumn{1}{c|}{$2.78\mathrm{e}{-05}$} & \multicolumn{1}{c|}{$0.001$}   & \multicolumn{1}{c|}{$0.0007$}  & \multicolumn{1}{c|}{$1.39\mathrm{e}{-05}$} & $0.005$              \\ \cline{2-10}
\multicolumn{1}{|c|}{}                               & CP   & \multicolumn{1}{c|}{5.29\%}                    & \multicolumn{1}{c|}{7.19\%}  & \multicolumn{1}{c|}{10.51\%} & \multicolumn{1}{c|}{$28.14\%$}             & \multicolumn{1}{c|}{$7.19\%$}  & \multicolumn{1}{c|}{$10.67\%$} & \multicolumn{1}{c|}{$29.41\%$}             & $24.99\%$             \\ \hline
\end{tabular}}
    \caption{1:1 reconstruction results. Mean errors and standard variances are in meters. CP is average compression percentage.}
    \label{tab:1:1 compression}
\end{table}

\begin{table}[]
\centering
\resizebox{\columnwidth}{!}{
\begin{tabular}{|c|ccc|ccc|ccc|}
\hline
\multirow{2}{*}{Dataset}                                       & \multicolumn{3}{c|}{Newer College}                                  & \multicolumn{3}{c|}{KITTI}                                          & \multicolumn{3}{c|}{INDOOR}                                         \\ \cline{2-10} 
                                                               & \multicolumn{1}{c|}{mean}   & \multicolumn{1}{c|}{std}    & CP      & \multicolumn{1}{c|}{mean}   & \multicolumn{1}{c|}{std}    & CP      & \multicolumn{1}{c|}{mean}   & \multicolumn{1}{c|}{std}    & CP      \\ \hline
\begin{tabular}[c]{@{}c@{}}CURL\\ (1:1 task only)\end{tabular} & \multicolumn{1}{c|}{0.0033} & \multicolumn{1}{c|}{0.0052} & 39.96\% & \multicolumn{1}{c|}{0.0031} & \multicolumn{1}{c|}{0.0046} & 36.53\% & \multicolumn{1}{c|}{0.0039} & \multicolumn{1}{c|}{0.0054} & 16.73\% \\ \hline
\end{tabular}
}
\caption{1:1 reconstruction result of CURL (fully tuned for 1:1 reconstruction).}
\label{tab:1:1 only}
\end{table}

\begin{figure}
    \centering
    \begin{subfigure}[b]{0.23\textwidth}
        \centering
        \includegraphics[width=\textwidth]{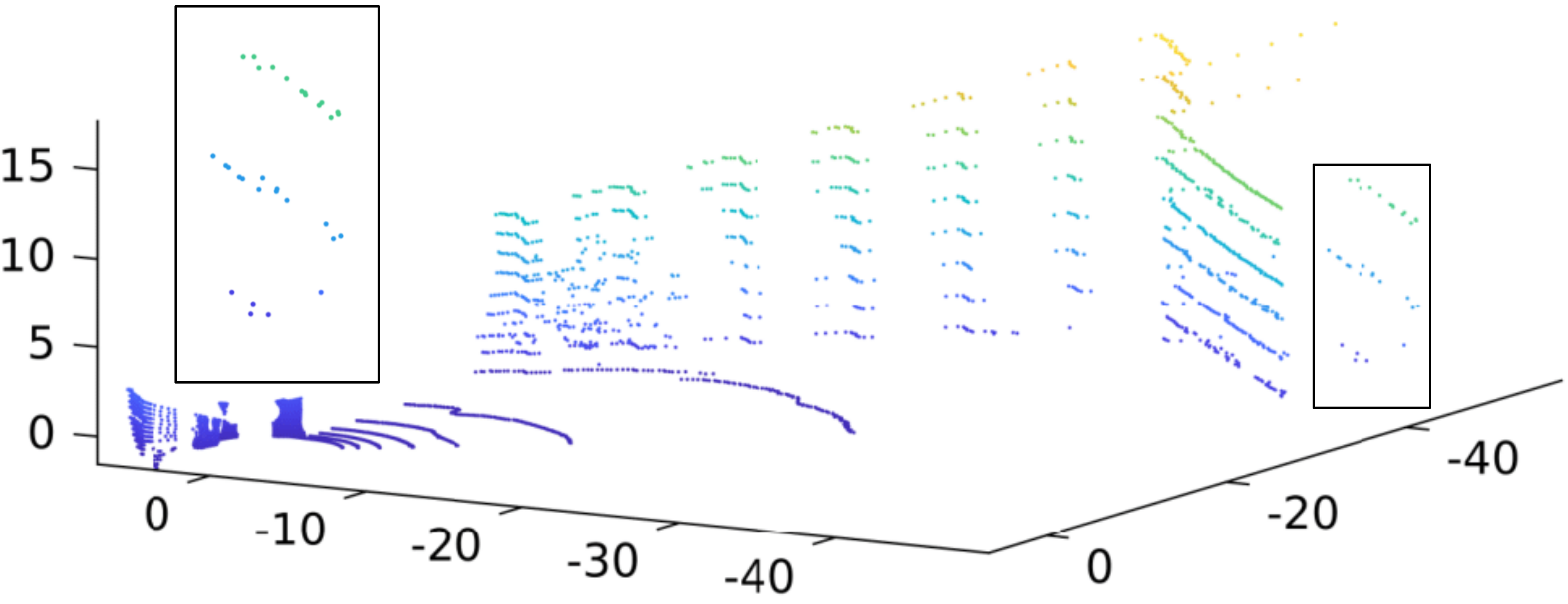}
        \caption{16-channel}
        \label{fig:16 to 64 LiDAR Newer College Dataset a}
    \end{subfigure}
    \hfill
    \begin{subfigure}[b]{0.23\textwidth}
        \centering
        \includegraphics[width=\textwidth]{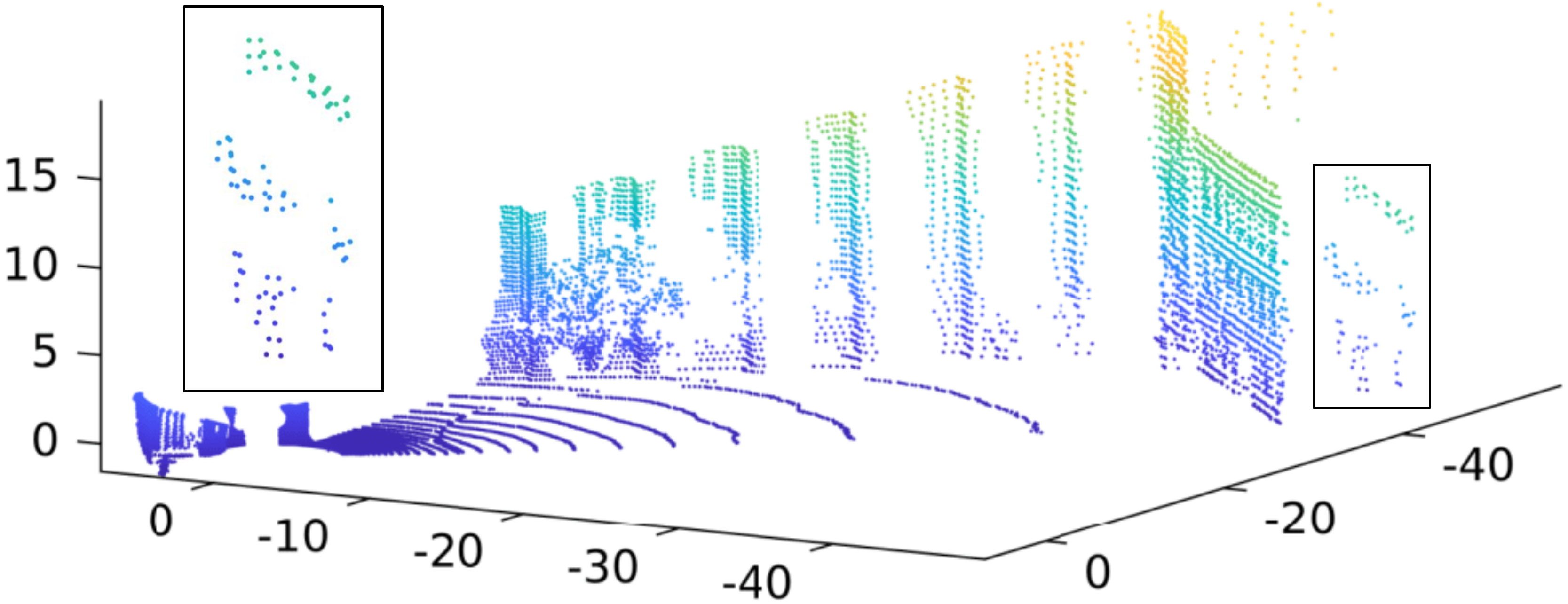}
        \caption{Ground truth}
        \label{fig:16 to 64 LiDAR Newer College Dataset b}
    \end{subfigure}
    \hfill
    \begin{subfigure}[b]{0.23\textwidth}
        \centering
        \includegraphics[width=\textwidth]{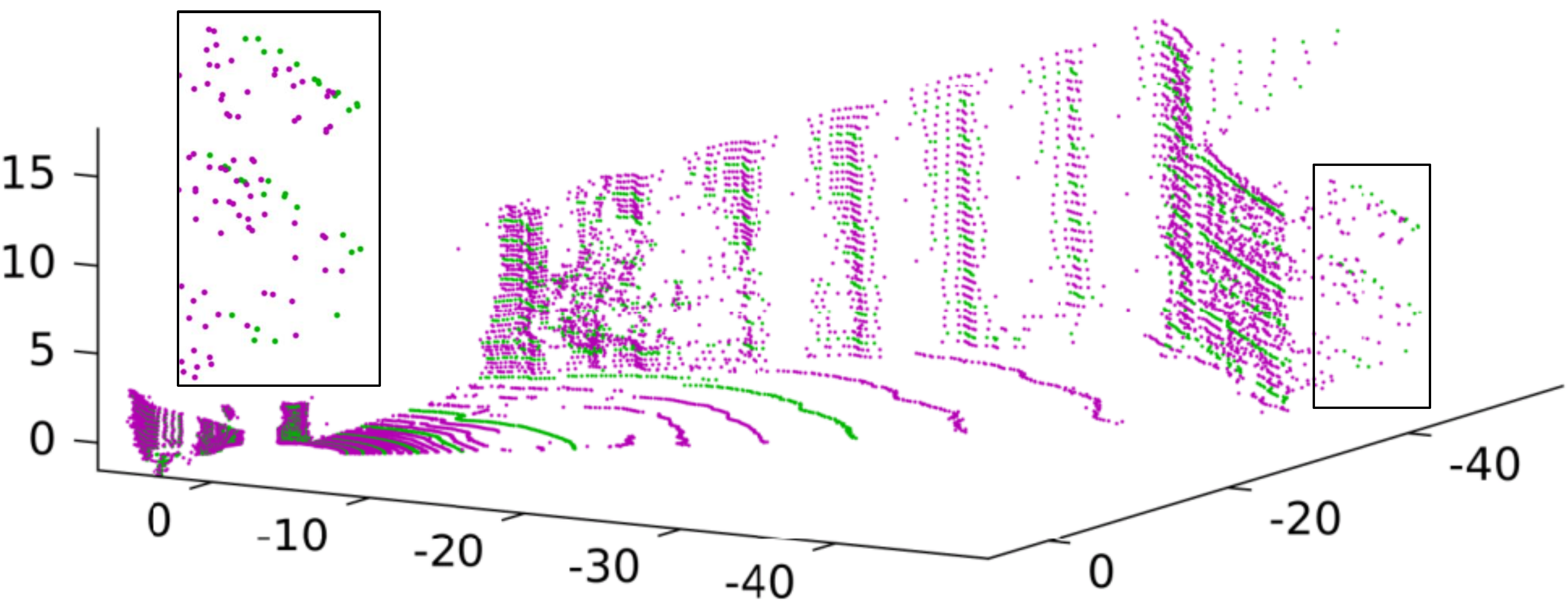}
     \caption{SuperRso}
        \label{fig:16 to 64 LiDAR Newer College Dataset c}
    \end{subfigure}
    \hfill
    \begin{subfigure}[b]{0.23\textwidth}
        \centering
        \includegraphics[width=\textwidth]{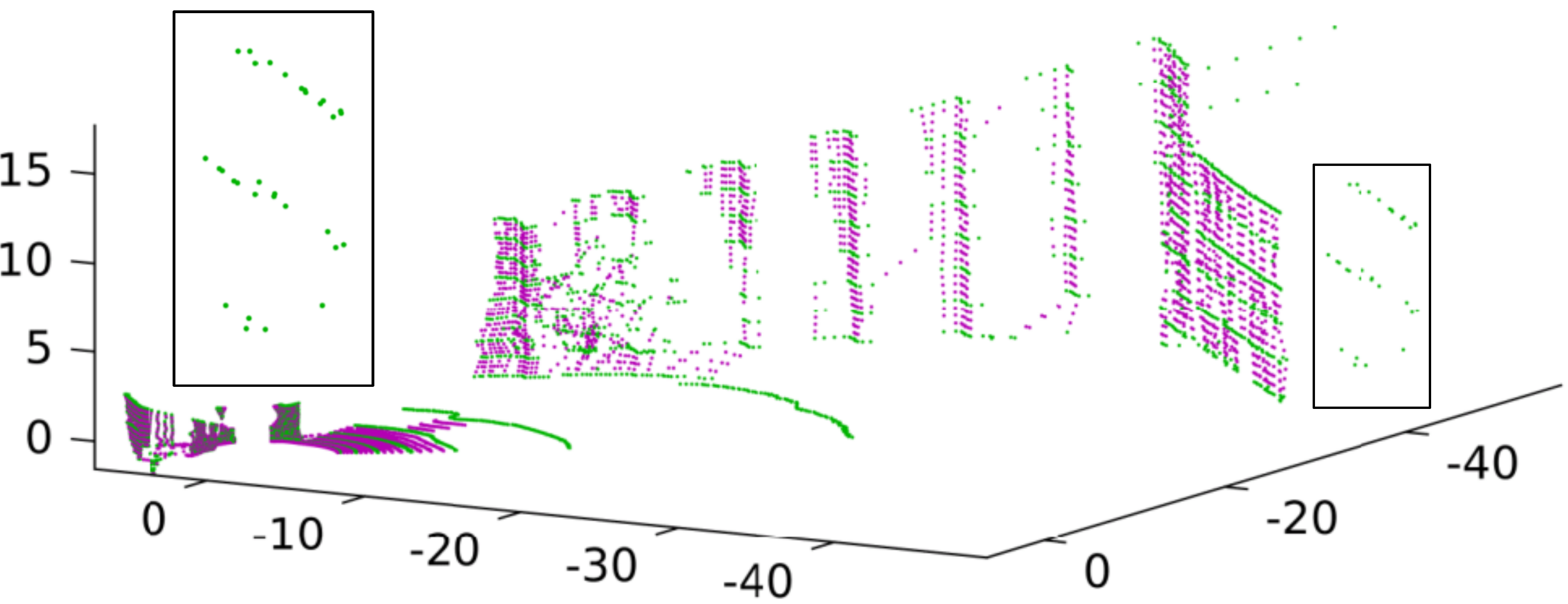}
         \caption{Ours}
        \label{fig:16 to 64 LiDAR Newer College Dataset d}
    \end{subfigure}
   \caption{A 16-to-64 channel upsampling example on Newer College dataset.}
   \label{fig:16 to 64 LiDAR Newer College Dataset}
\end{figure}

TABLE \ref{tab:1:1 compression} shows the results of baseline, PCL, and CURL on the Newer College, KITTI, and INDOOR datasets. It can be seen that the baseline method achieves the best compression percentage yet with the lowest reconstruction accuracy. PCL has superior accuracy and compression percentage performance, although it cannot increase density or provide continuous reconstruction. CURL's performance is inferior to PCL, but it is able to balance the accuracy and compression percentage reasonably well. Fig. \ref{fig:all datasets} provides cumulative distribution function (CDF) of CURL's reconstruction errors and compression percentage in detail.
Some 1:1 reconstruction examples are given in Fig. \ref{fig:1:1 reconstruction for three datasets}. 
Since the same CURL derived here will also be used for continuous reconstruction (our main focus) in Section \ref{sec:EvaluationCR} and it is preferably to avoid overfitting on a single task, its 1:1 reconstruction results in TABLE \ref{tab:1:1 compression} are not the best. However, if CURL is tuned mostly for the 1:1 reconstruction task, it can achieve comparable performance to PCL, as shown in the TABLE \ref{tab:1:1 only}.

\begin{figure}
    \centering
    \begin{subfigure}[b]{0.48\linewidth}
        \centering
        \includegraphics[height=1.2in,width=\linewidth]{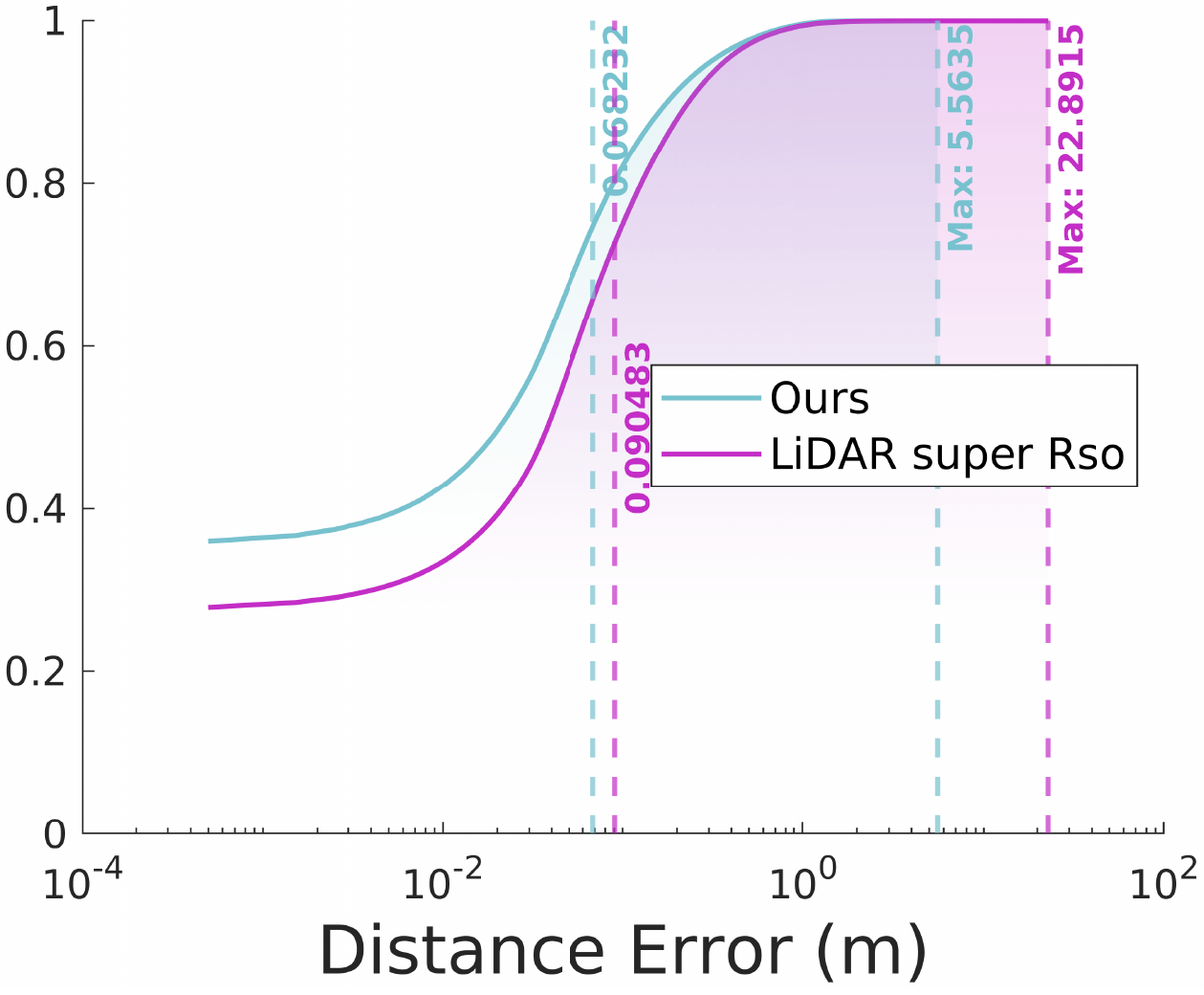}
        \caption{SuperRso dataset}
        \label{fig:up sampling experiment a}
    \end{subfigure}
    \hfill
    \begin{subfigure}[b]{0.48\linewidth}
        \centering
        \includegraphics[height=1.2in,width=\linewidth]{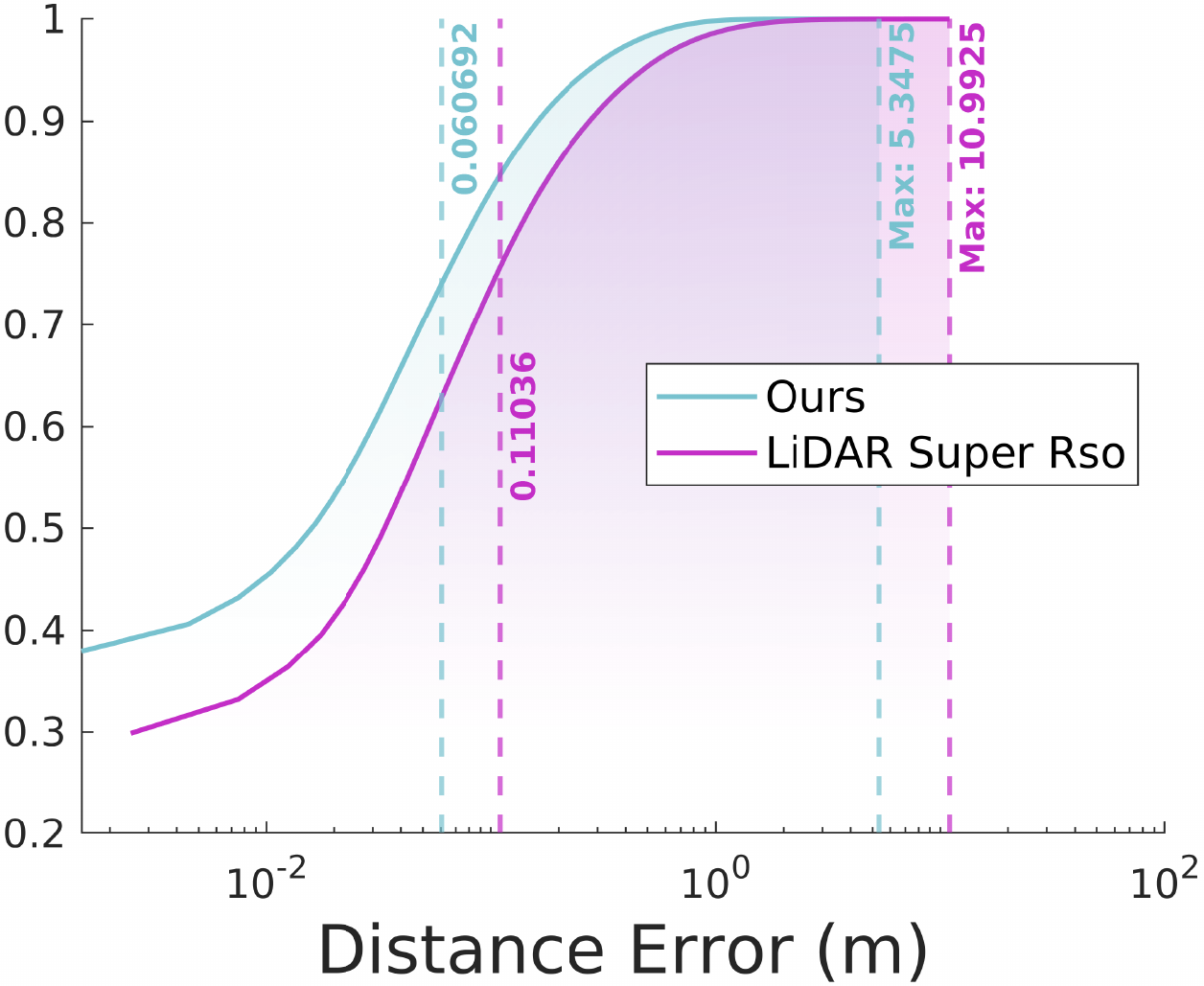}
        \caption{Newer College dataset}
        \label{fig:up sampling experiment b}
    \end{subfigure}
       \caption{Cumulative distributions of our upsampling and SuperRso methods on reconstruction errors.}
       \label{fig:up sampling experiment}
\end{figure}

\begin{figure}
    \centering
    \begin{subfigure}[b]{0.23\textwidth}
        \centering
        \includegraphics[width=\textwidth]{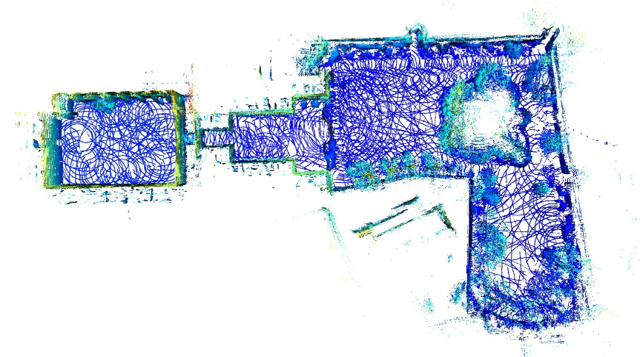}
        \caption{16-channel}
        \label{fig:16 to 64 LiDAR SuperRso dataset mapping a}
    \end{subfigure}
    \hfill
    \begin{subfigure}[b]{0.23\textwidth}
        \centering
        \includegraphics[width=\textwidth]{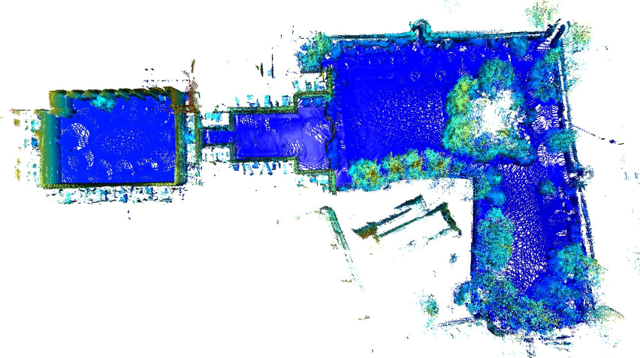}
        \caption{Ground truth}
        \label{fig:16 to 64 LiDAR Newer College Dataset mapping b}
    \end{subfigure}
    \hfill
    \begin{subfigure}[b]{0.23\textwidth}
        \centering
        \includegraphics[width=\textwidth]{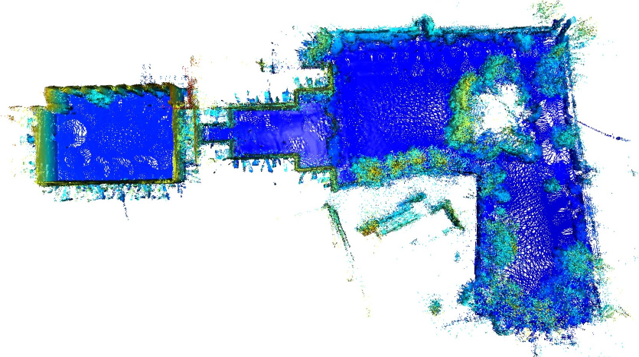}
     \caption{SuperRso}
        \label{fig:16 to 64 LiDAR Newer College Dataset mapping c}
    \end{subfigure}
    \hfill
    \begin{subfigure}[b]{0.23\textwidth}
        \centering
        \includegraphics[width=\textwidth]{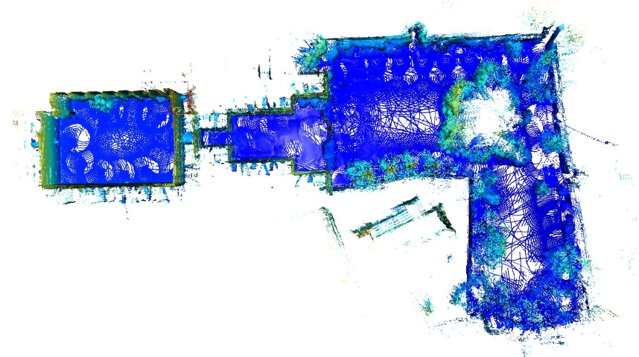}
     \caption{Ours}
        \label{fig:16 to 64 LiDAR Newer College Dataset mapping d}
    \end{subfigure}
    \hfill
   \caption{Mapping results. SuperRso and our method use their 16-to-64 upsampled scans.
}
   \label{fig:16 to 64 LiDAR Newer College Dataset mapping}
\end{figure}
\begin{figure}
   \centering
   \includegraphics[width=\linewidth]{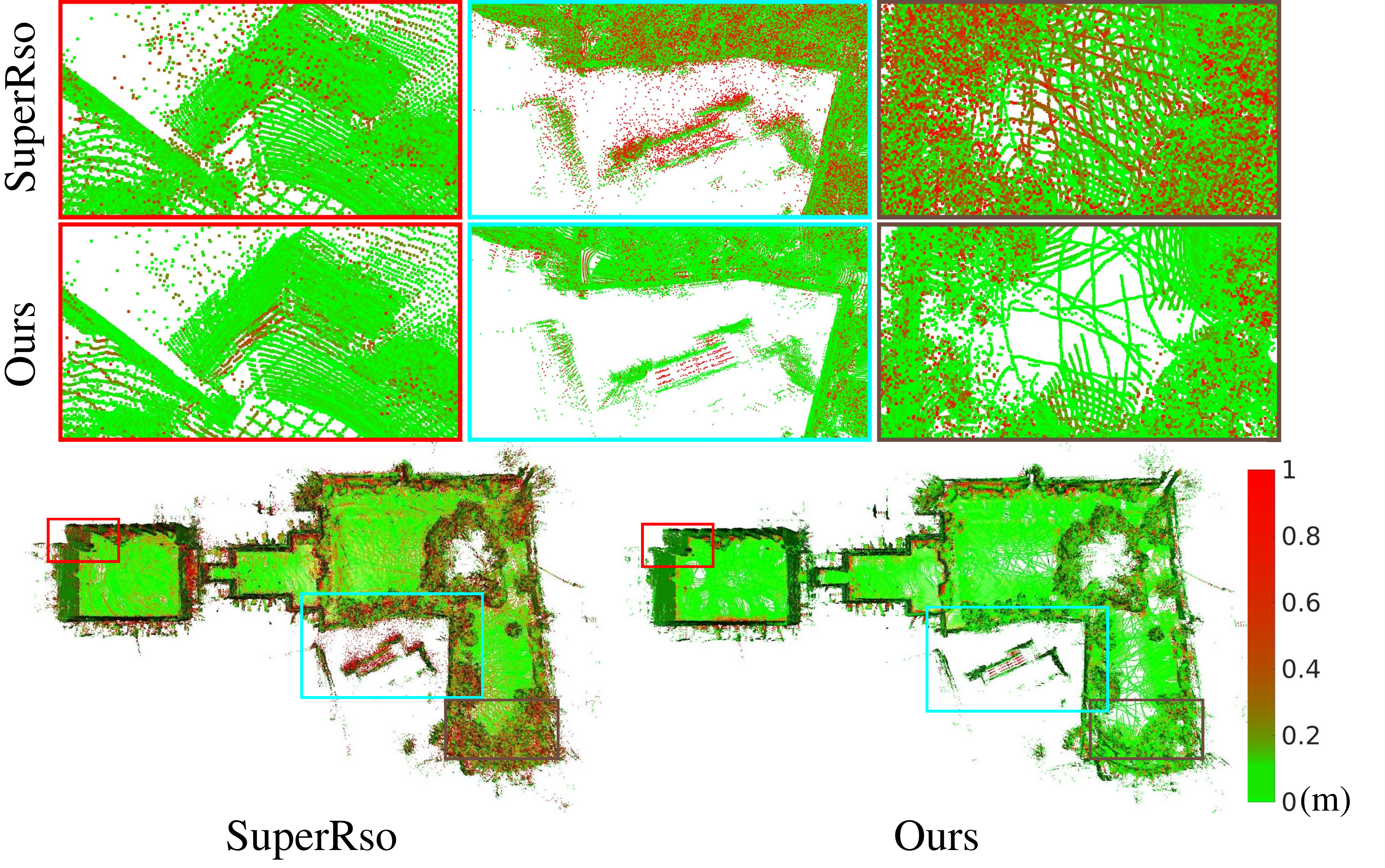}
   \caption{Error maps compared with ground truth map. }
\label{fig:up sampling error figure}
\end{figure}

\subsection{Evaluation on Upsampling}
This evaluation compares our upsampling algorithm with the LiDAR super-resolution method SuperRso \cite{shan2020simulation}, which is a recent open-source system that for LiDAR point cloud upsampling.
Since the available pre-trained model of SuperRso is to increase LiDAR scan from 16-channel to 64-channel, we use the dataset provided by SuperRso and the Newer College datasets for evaluation by sampling 16-channel LiDAR scans from Ouster's raw 64-channel scans.
An example on the Newer College dataset is shown in Fig. \ref{fig:16 to 64 LiDAR Newer College Dataset}.
Compared with SuperRso, our method predicts fewer points at the areas that have sparse points in the original 16-channel and conservatively recovers a point cloud that preserves similar geometric structures in the ground truth point cloud.
Fig. \ref{fig:up sampling experiment} provides detailed CDF of the reconstruction errors of these two methods.
Although our method is more conservative and generates relatively fewer points than SuperRso, its precision tends to be higher. This is important for our CURL methods as precision is the priority for upsampling, whose point cloud forms the basis to determine the encoding quality. Moreover, its continuous reconstruction can always significantly increase density later.
To further understand the quality of upsampled scans, all scans are compounded into a consistent 3D map using the ground truth poses in the Newer College dataset, as shown in Fig. \ref{fig:16 to 64 LiDAR Newer College Dataset mapping} and Fig. \ref{fig:up sampling error figure}. It is evident that the mapping results of using SuperRso and our method have much higher density compared with the 16-channel results, and the geometric structures are well maintained, although SuperRso produces a bit higher number of noise points.
\begin{table}[]
\resizebox{\columnwidth}{!}{
\centering
\begin{tabular}{|c|cc|cc|cc|cc|c|}
\hline
\multirow{2}{*}{Dataset}       & \multicolumn{2}{c|}{\multirow{2}{*}{Method}}                          & \multicolumn{2}{c|}{$R_{row}=2$, $R_{col}=1$}         & \multicolumn{2}{c|}{$R_{row}=4$, $R_{col}=1$}         & \multicolumn{2}{c|}{$R_{row}=8$, $R_{col}=1$}         & \multirow{2}{*}{CP} \\ \cline{4-9}
                               & \multicolumn{2}{c|}{}                                                 & \multicolumn{1}{c|}{mean}     & std                   & \multicolumn{1}{c|}{mean}     & std                   & \multicolumn{1}{c|}{mean}     & std                   &                     \\ \hline
\multirow{4}{*}{\shortstack{Newer\\College}} & \multicolumn{2}{c|}{Baseline}                                         & \multicolumn{1}{c|}{105.6024} & $1.39\mathrm{e}{+05}$ & \multicolumn{1}{c|}{175.6662} & $1.93\mathrm{e}{+05}$ & \multicolumn{1}{c|}{212.7353} & $2.20\mathrm{e}{+05}$ & 7.94\%              \\ \cline{2-10}
                               & \multicolumn{1}{c|}{\multirow{3}{*}{Ours}} & $S_{row}=1$, $S_{col}=1$ & \multicolumn{1}{c|}{42.9162}  & $4.06\mathrm{e}{+06}$ & \multicolumn{1}{c|}{56.3463}  & $5.29\mathrm{e}{+04}$ & \multicolumn{1}{c|}{64.8137}  & $6.10\mathrm{e}{+04}$ & 27.87\%             \\ \cline{3-10}
                               & \multicolumn{1}{c|}{}                      & $S_{row}=2$, $S_{col}=1$ & \multicolumn{1}{c|}{0.0628}   & $0.0450$              & \multicolumn{1}{c|}{0.1303}   & 117.0900              & \multicolumn{1}{c|}{0.1509}   & $124.0000$            & 26.90\%             \\ \cline{3-10}
                               & \multicolumn{1}{c|}{}                      & $S_{row}=2$, $S_{col}=2$ & \multicolumn{1}{c|}{0.0619}   & $0.0430$              & \multicolumn{1}{c|}{0.0671}   & 10.6600               & \multicolumn{1}{c|}{0.0684}   & $11.3400$             & 24.71\%             \\ \hline
\multirow{4}{*}{INDOOR}        & \multicolumn{2}{c|}{Baseline}                                         & \multicolumn{1}{c|}{0.0234}   & 0.2490                & \multicolumn{1}{c|}{0.0249}   & 0.7285                & \multicolumn{1}{c|}{0.0258}   & 0.8260                & 5.29\%              \\ \cline{2-10}
                               & \multicolumn{1}{c|}{\multirow{3}{*}{Ours}} & $S_{row}=1$, $S_{col}=1$ & \multicolumn{1}{c|}{0.6418}   & $2.09\mathrm{e}{+02}$ & \multicolumn{1}{c|}{0.7953}   & $2.16\mathrm{e}{+02}$ & \multicolumn{1}{c|}{0.8353}   & $2.16\mathrm{e}{+02}$ & 32.30\%             \\ \cline{3-10}
                               & \multicolumn{1}{c|}{}                      & $S_{row}=2$, $S_{col}=1$ & \multicolumn{1}{c|}{0.0042}   & 0.0055                & \multicolumn{1}{c|}{0.0075}   & 2.8265                & \multicolumn{1}{c|}{0.0084}   & 2.9849                & 27.90\%             \\ \cline{3-10}
                               & \multicolumn{1}{c|}{}                      & $S_{row}=2$, $S_{col}=2$ & \multicolumn{1}{c|}{0.0041}   & 0.0048                & \multicolumn{1}{c|}{0.0040}   & 0.1106                & \multicolumn{1}{c|}{0.0040}   & 0.1167                & 24.99\%             \\ \hline
\end{tabular}}
\caption{Continuous reconstruction for $P_r=4$ with different sampling rates. Sampling rate of baseline method is fixed as $S_{row}=S_{col}=2$. Mean errors and standard variances are in meter.}
    \label{tab:continuous reconstruction smapling rate}
\end{table}

\begin{figure}
    \centering
    \begin{subfigure}[b]{0.48\linewidth}
        \centering
            \includegraphics[width=\linewidth]{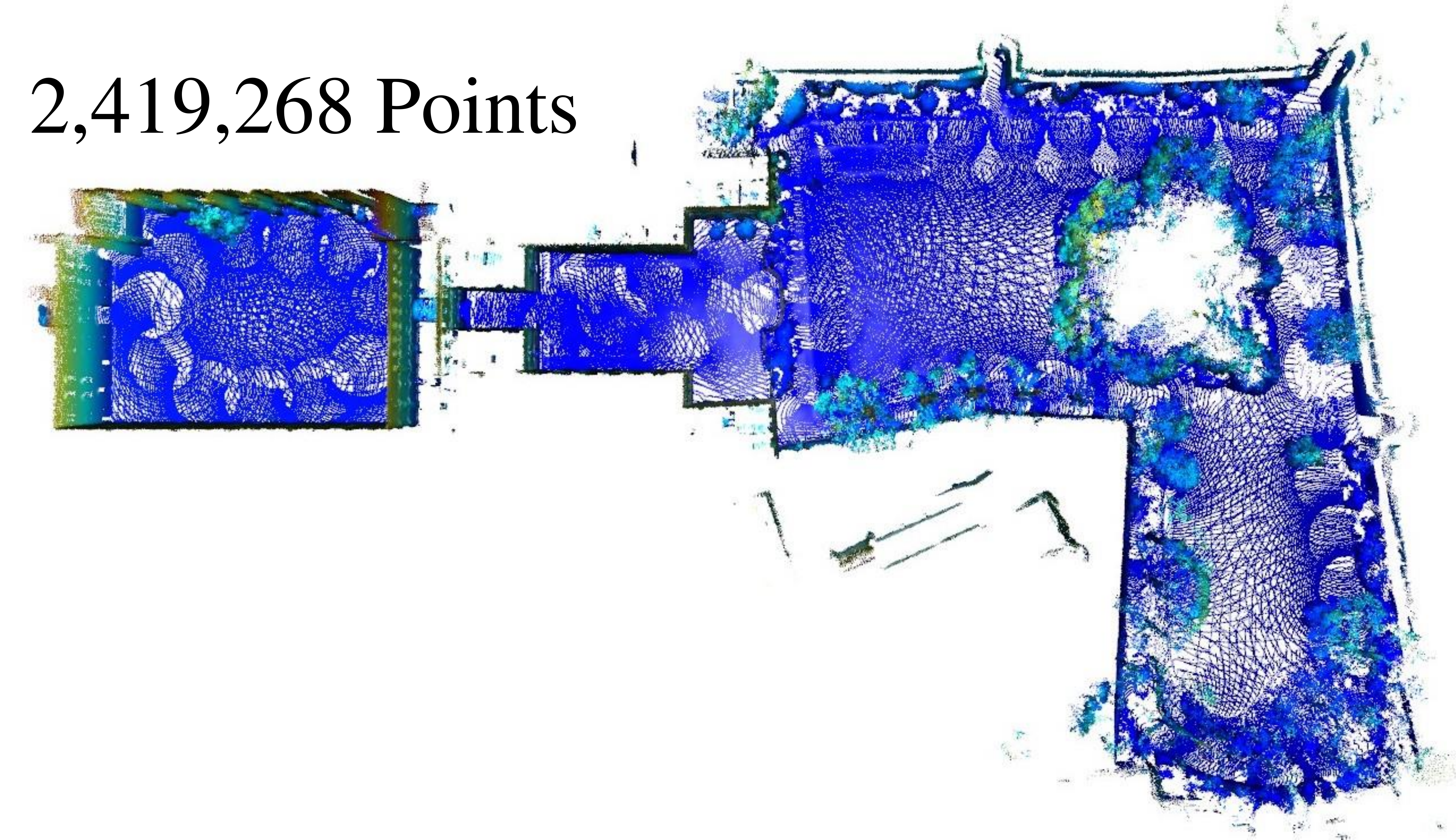}
            \caption{Map using 64-channel scans}
            \label{fig:continuous reconstruction with gt a}
        \end{subfigure}
        \hfill
        \begin{subfigure}[b]{0.48\linewidth}
            \centering
            \includegraphics[width=\linewidth]{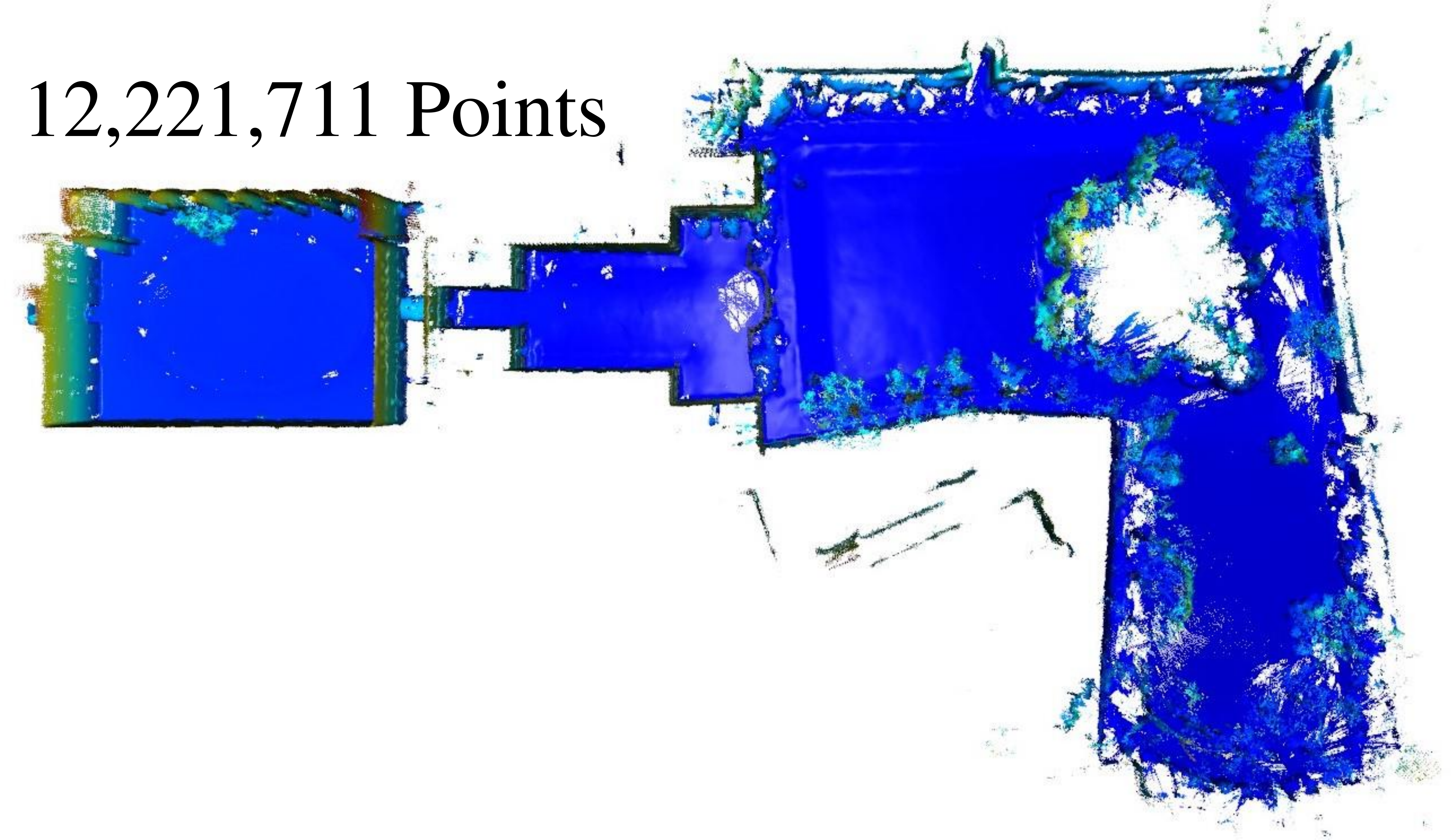}
            \caption{Map using CURL scans}
            \label{fig:continuous reconstruction with gt b}
        \end{subfigure}
           \caption{Mapping on the Newer College dataset. (a) Map using the original 64-channel scans from the Ouster LiDAR. (b) Map using the scans reconstructed by CURL ($R_{row}=8$). }
           \label{fig:continuous reconstruction with gt}
   \end{figure}
   \begin{figure}
       \centering
       \includegraphics[width=\linewidth]{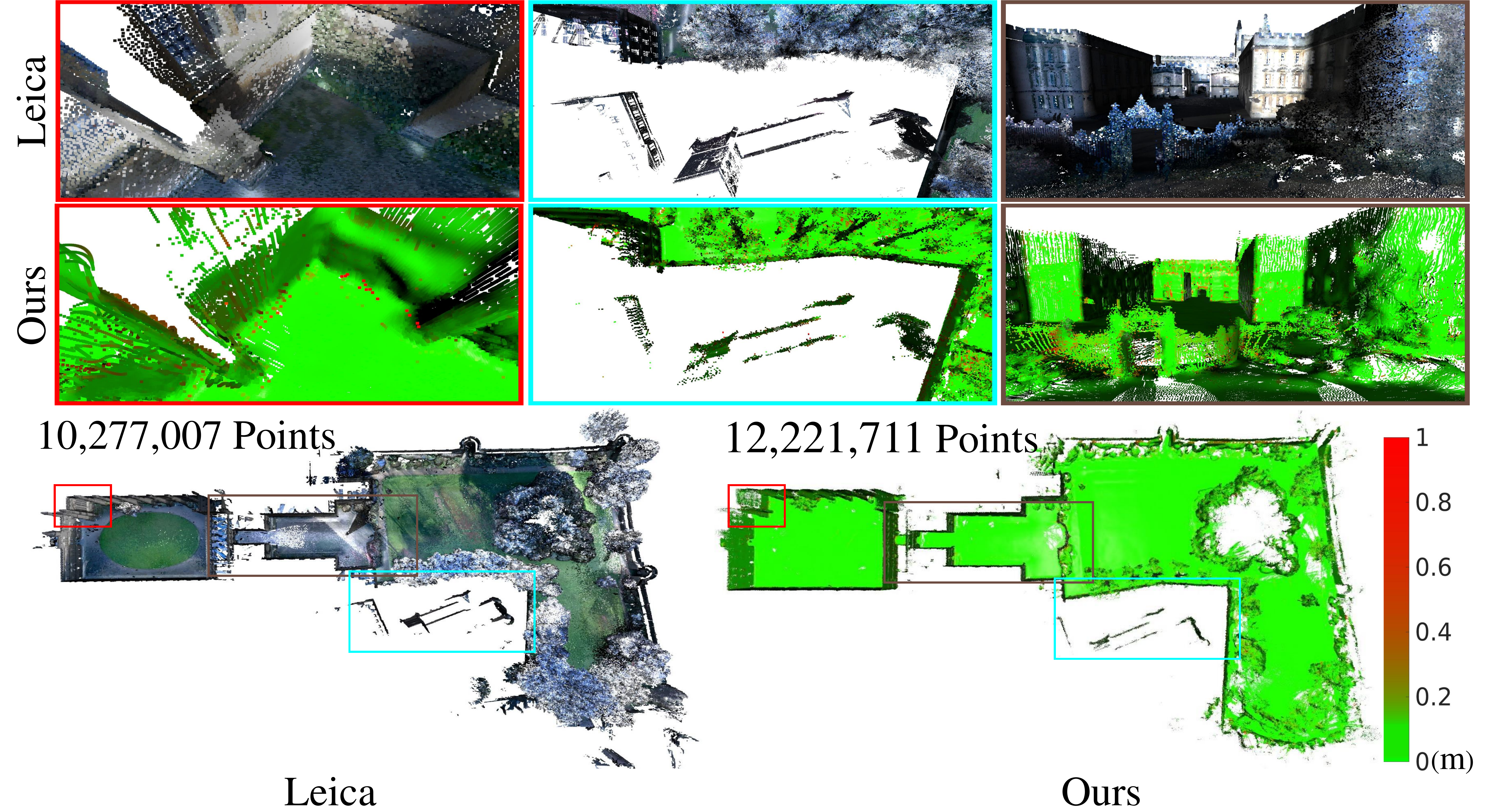}
       \caption{CURL continuous reconstruction map in Fig. \ref{fig:continuous reconstruction with gt b} compared with the Leica ground truth map.}
   \label{fig:leica vs continuous}
   \end{figure}

\subsection{Evaluation on Continuous Reconstruction}\label{sec:EvaluationCR}

Continuous reconstruction is to increase the number of points in a point cloud, reaching a higher density than its original one. We show some qualitative examples first before giving detailed quantitative evaluations.




\subsubsection{Qualitative Evaluation}
\begin{figure}
    \centering
    \begin{subfigure}[b]{0.32\linewidth}
        \centering
        \includegraphics[width=\linewidth]{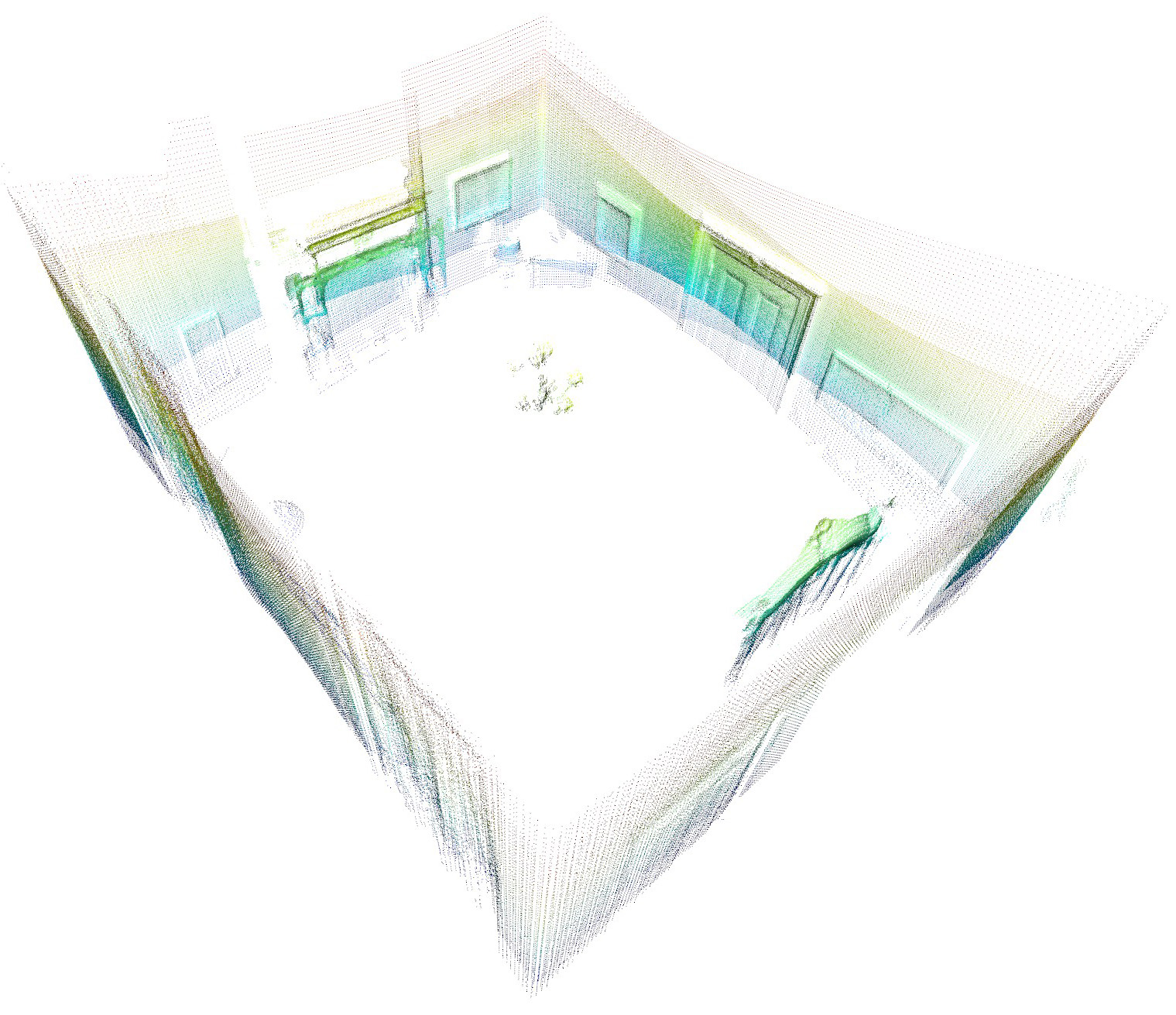}
        \caption{64-channel map}
    \end{subfigure}
    \begin{subfigure}[b]{0.32\linewidth}
        \centering
        \includegraphics[width=\linewidth]{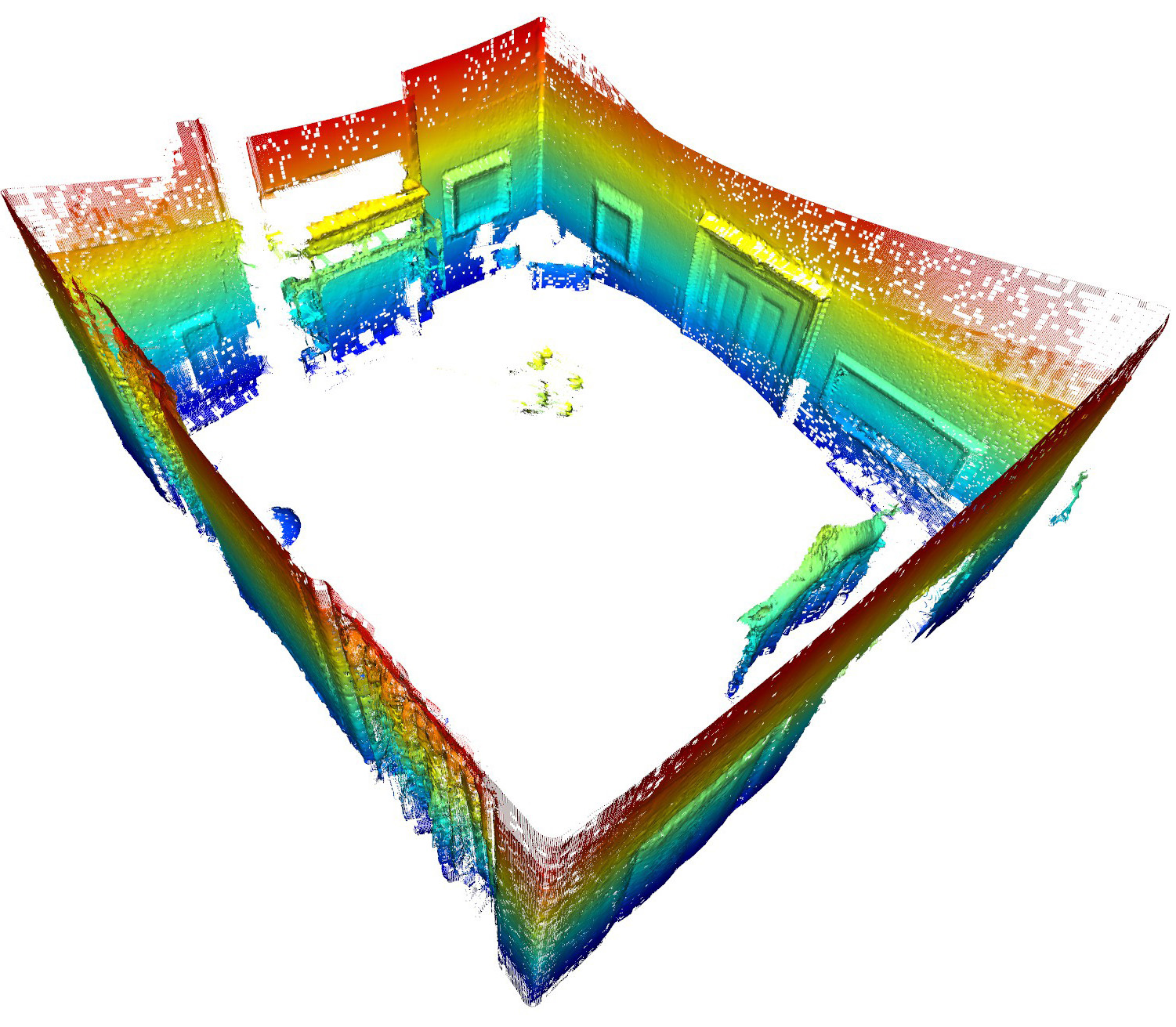}
        \caption{CURL map}
    \end{subfigure}
    \begin{subfigure}[b]{0.32\linewidth}
        \centering
        \includegraphics[width=\linewidth]{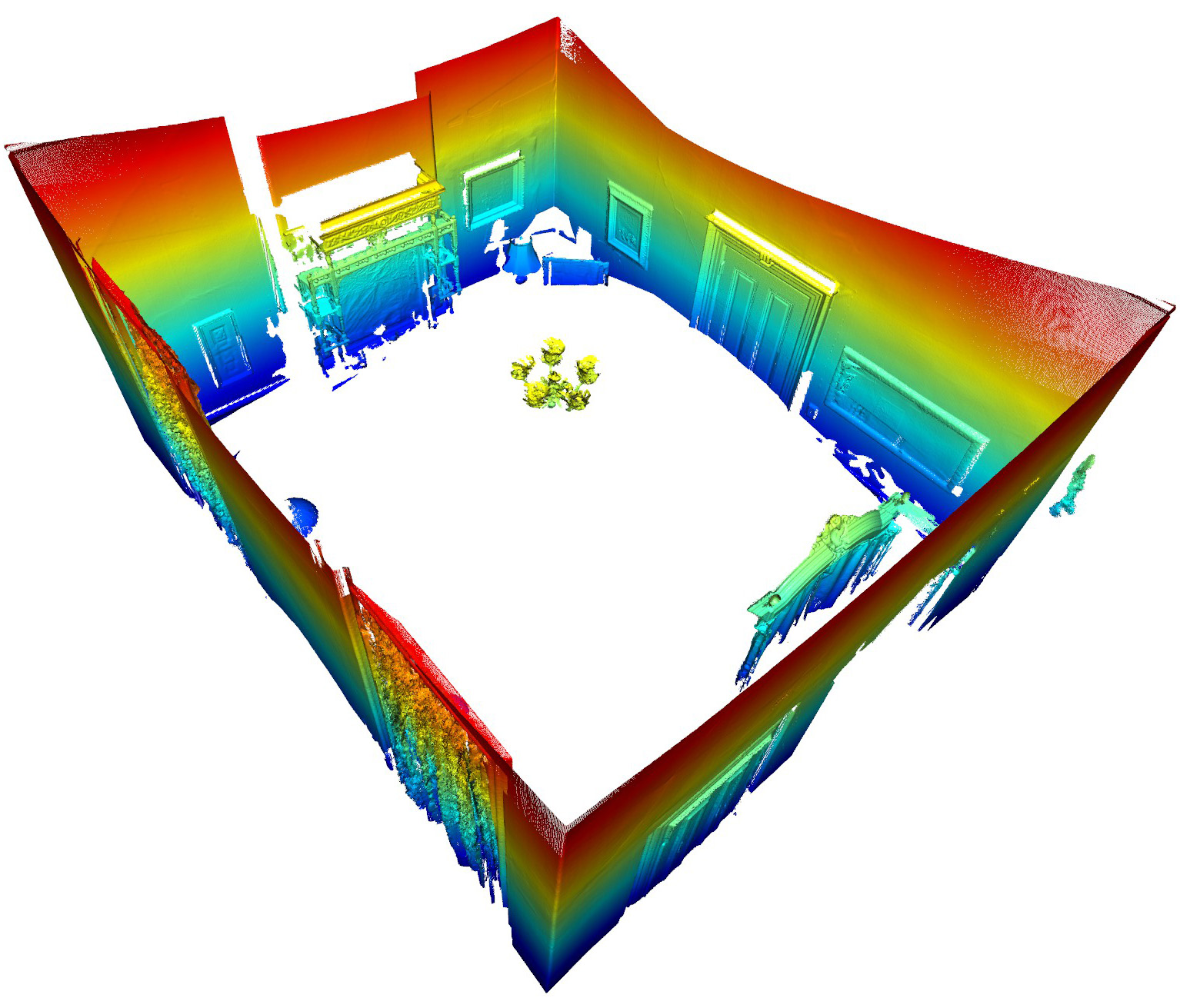}
        \caption{Ground truth map}
    \end{subfigure}
    \hfill
    \begin{subfigure}[b]{0.32\linewidth}
        \centering
        \includegraphics[width=\linewidth]{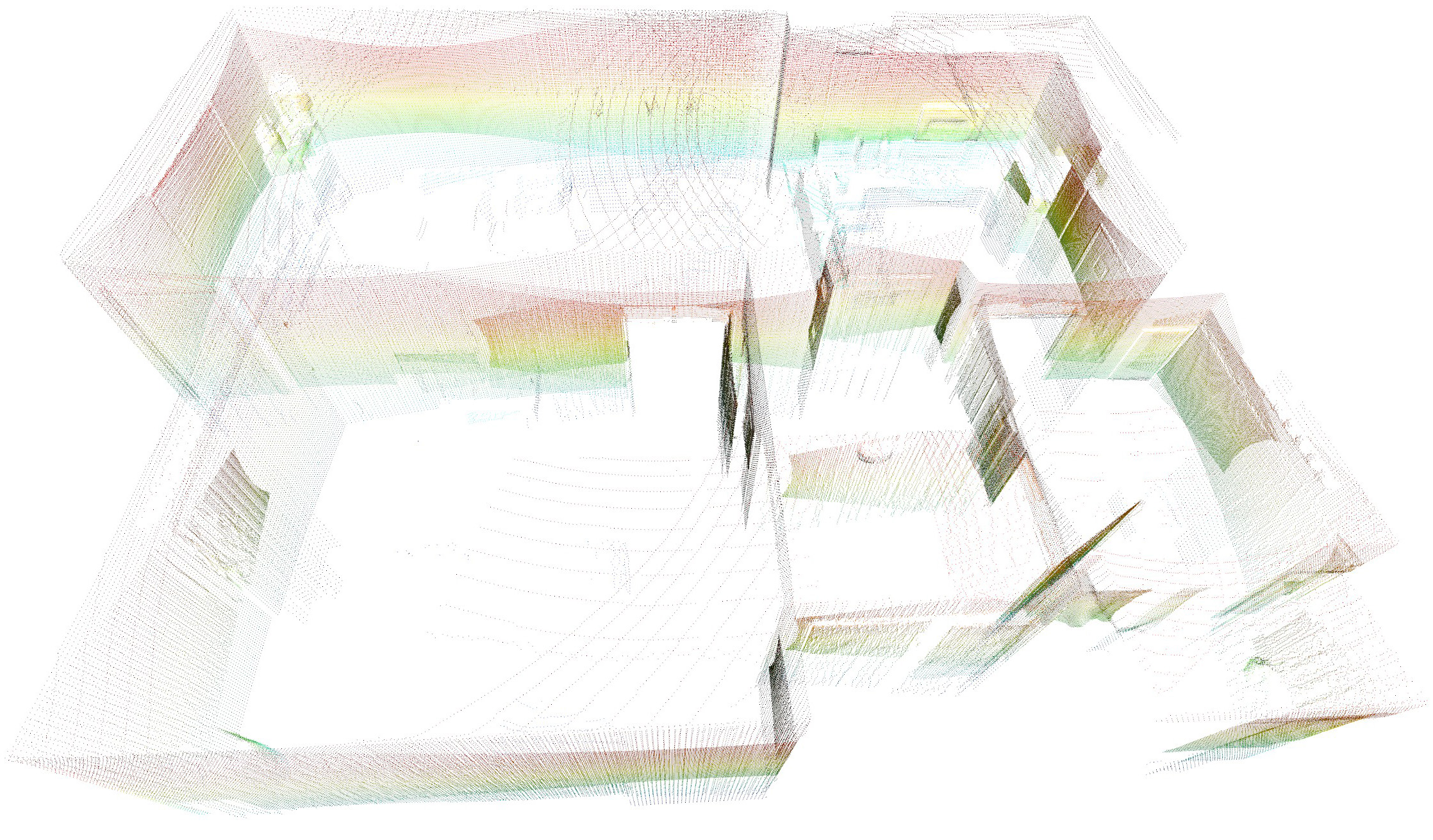}
        \caption{64-channel map}
    \end{subfigure}
    \begin{subfigure}[b]{0.32\linewidth}
        \centering
        \includegraphics[width=\linewidth]{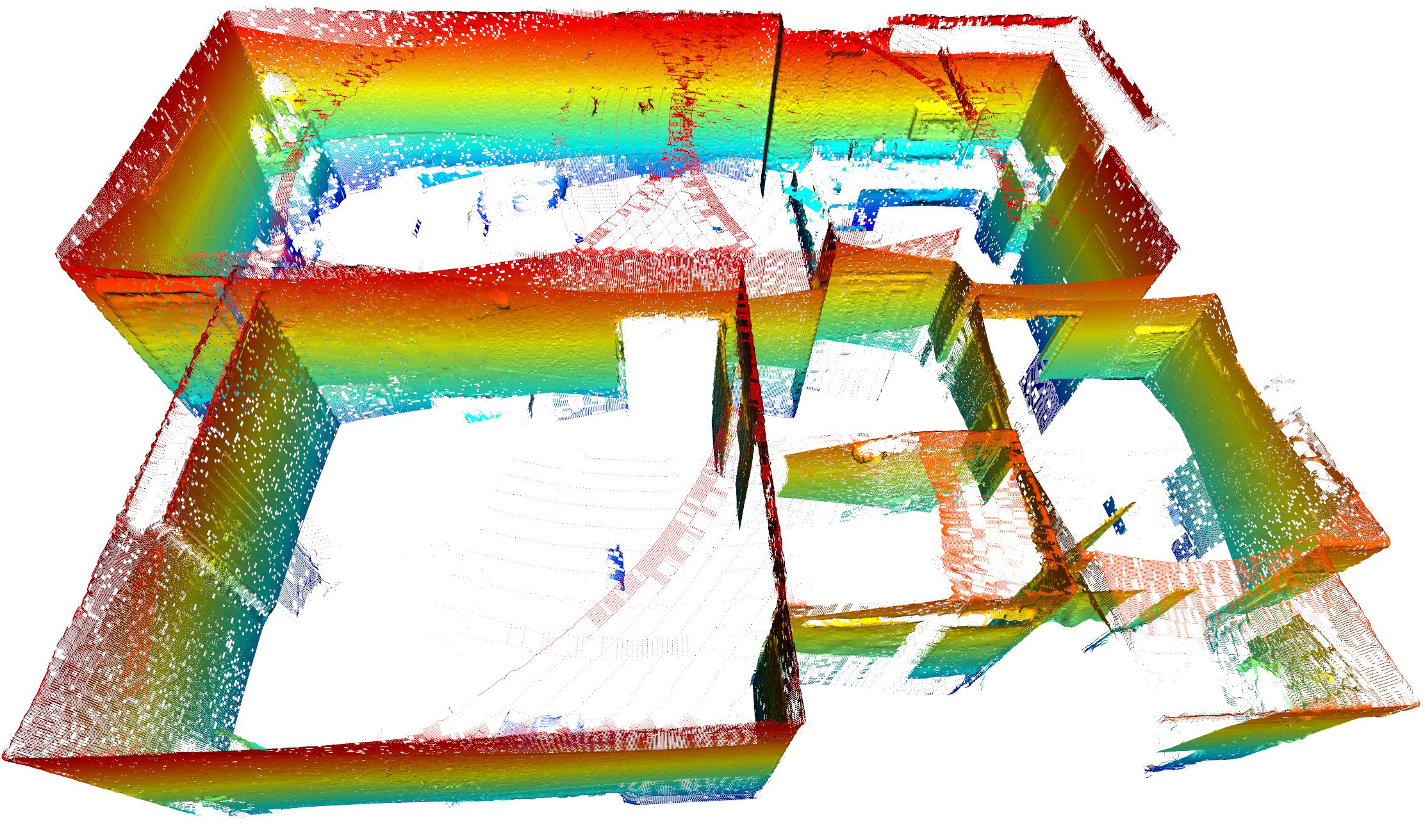}
        \caption{CURL map}
    \end{subfigure}
    \begin{subfigure}[b]{0.32\linewidth}
        \centering
        \includegraphics[width=\linewidth]{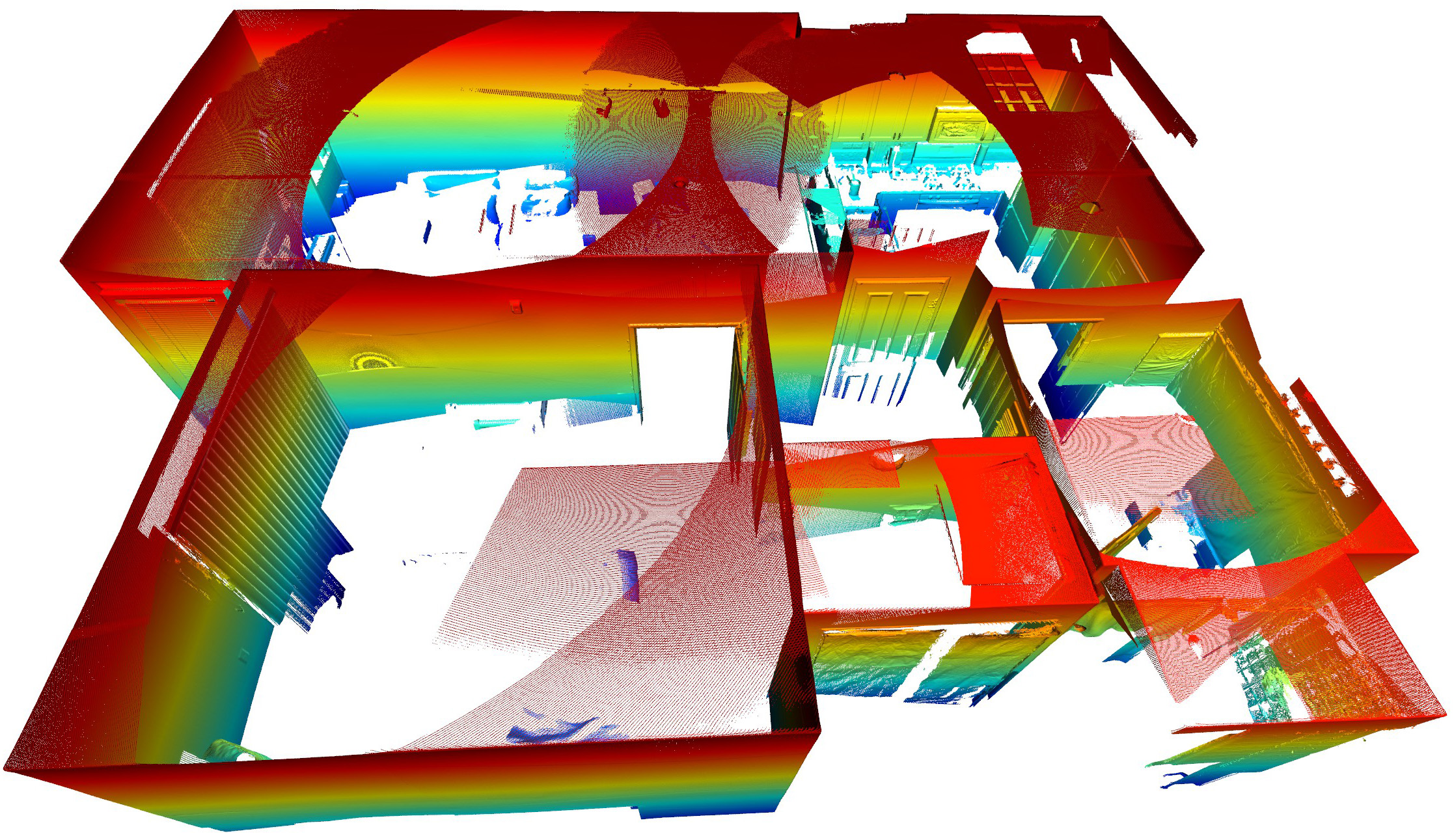}
        \caption{Ground truth map}
    \end{subfigure}
   \caption{Two sets of mapping results using INDOOR dataset. CURL map is built from continuous reconstruction using the 64-channel map in first column. Ground truth maps are built by a FARO X330 static scanner.}
   \label{fig:CR_Indoor}
\end{figure}

\begin{figure}
    \centering
    \includegraphics[width=\linewidth]{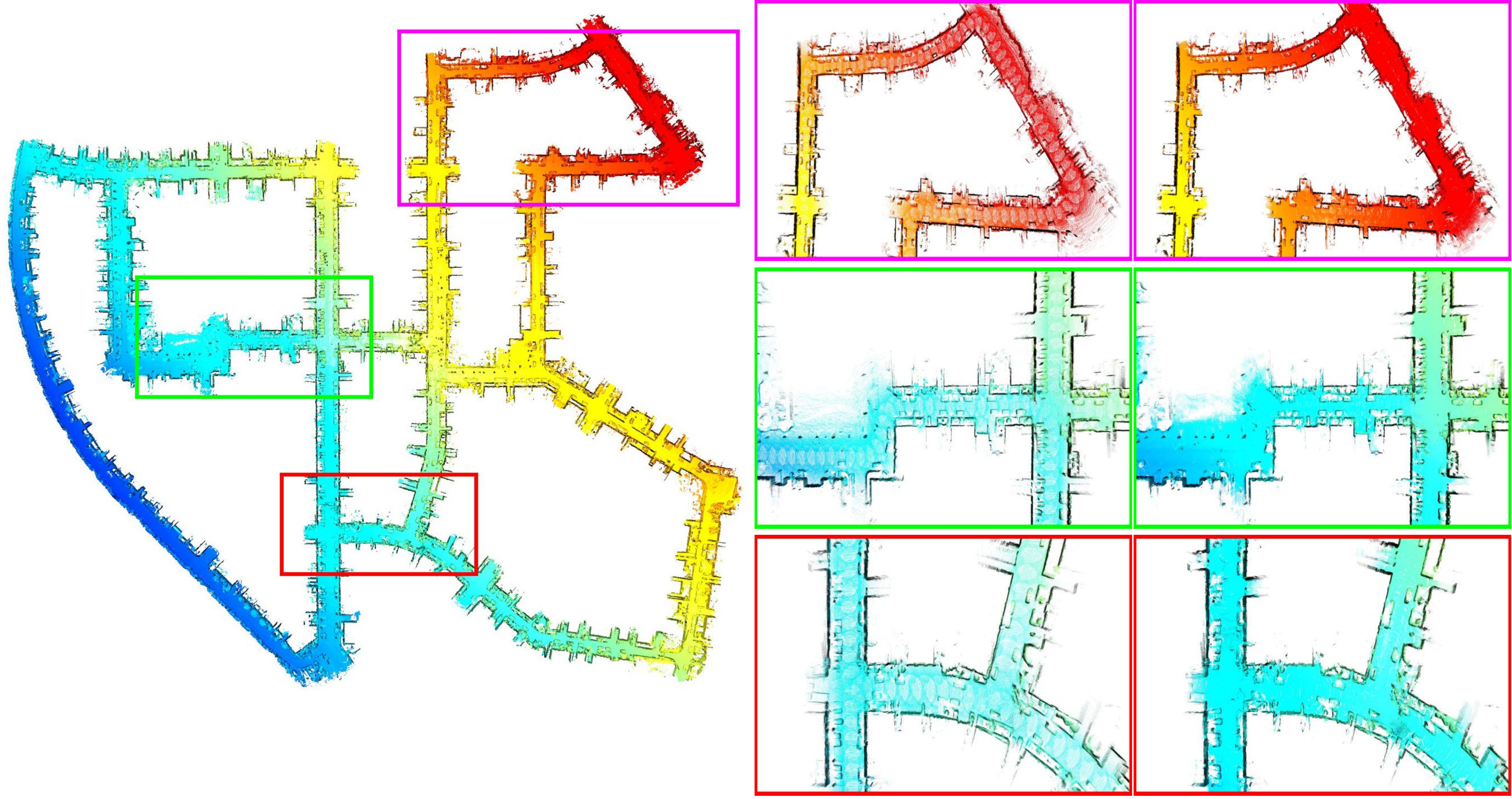}
    \caption{Comparison of maps built using the KITTI Sequence 0 dataset. Left: map of original 64-channel LiDAR scans with about 19 million points (430.25MB). Right: map of the CURL scans ($R_{row}=8$ and $R_{col}=1$ continuous reconstruction) with about 98 million points (CURL size: 71.25MB).}
\label{fig:KITTI continuous reconstruction}
\end{figure}

Fig. \ref{fig:continuous reconstruction with gt} shows the maps using the original 64-channel scans captured from the Ouster LiDAR and the scans continuously reconstructed from CURL. We can see the CURL map is much denser with over $5\times$ number of points. But the binary size of the CURL used to reconstruct the map in Fig. \ref{fig:continuous reconstruction with gt b} only takes about $18.5\%$ of that of the point cloud map in Fig. \ref{fig:continuous reconstruction with gt a} (9.61MB vs. 51.93MB).
The errors of this CURL map compared with the Lecia ground truth map are described in Fig. \ref{fig:leica vs continuous}. This validates that the proposed CURL method is able to achieve reasonable accuracy comparing with this survey-grade Lecia scanner.
Fig. \ref{fig:CR_Indoor} shows extra examples of the continuous reconstruction results in indoor scenes using the INDOOR dataset. We can see some fine details of the wall arts, doors, etc., successfully recovered through the CURL continuous reconstruction.
In order to verify the performance in outdoor street environments, Fig. \ref{fig:KITTI continuous reconstruction} shows the continuous reconstruction results using the KITTI dataset. Since the KITTI dataset does not have a complete ground truth map, we compare the CURL map with the map built using the 64-channel LiDAR scans. It can be seen that the CURL method operates well in the outdoor street scene.

\subsubsection{Quantitative Evaluation}
To quantitatively evaluate the performance of continuous reconstruction, a denser and more accurate point cloud map is needed. Therefore, we use the Newer College and INDOOR datasets which have high-quality point clouds captured using survey-grade LiDAR scanners (Leica BLK360 and FARO 3D X330 scanners, respectively) as ground truth maps.
From TABLE \ref{tab:continuous reconstruction smapling rate}, we can see that with the higher sampling rates, the errors reduce dramatically. In contrast, even a high sampling rate is utilized for the baseline method, its reconstruction error remains big. This verifies the importance of the patching and adaptive refinement.

\subsection{Computation Time}
\begin{table}[]
\centering
\resizebox{0.5\textwidth}{!}{
\begin{tabular}{|c|c|c|c|}
\hline
                                      Modules                          & Extraction Time & Reconstruction Time & Total Time \\ \hline
\begin{tabular}[c]{@{}c@{}}CURL\\ (1:1 task only)\end{tabular} & 0.0444s         & 0.0073s              & 0.0517s    \\ \hline
\begin{tabular}[c]{@{}c@{}}PCL\\ (Oline Low-Res)\end{tabular}  & 0.0175s         & 0.0257s             & 0.0432s    \\ \hline
\end{tabular}}
\caption{Average computation time on KITTI Sequence 0.}
\label{tab:computation time 1:1}
\end{table}

\begin{table}[]
\resizebox{0.5\textwidth}{!}{
\begin{tabular}{|c|cc|c|cc|c|}
\hline
Modules               & \multicolumn{2}{c|}{Upsampling Time}                                                       & Extraction Time          & \multicolumn{2}{c|}{Reconstruction Time}               & Total Time \\ \hline
\multirow{4}{*}{CURL} & \multicolumn{1}{c|}{\multirow{4}{*}{$S_{row} = 2$ $S_{col}=2$}} & \multirow{4}{*}{0.3441s} & \multirow{4}{*}{0.0707s} & \multicolumn{1}{c|}{$R_{row}=2$ $R_{col}=1$} & 0.0079s & 0.4227s    \\ \cline{5-7} 
                      & \multicolumn{1}{c|}{}                                           &                          &                          & \multicolumn{1}{c|}{$R_{row}=4$ $R_{col}=1$} & 0.0152s & 0.4300s    \\ \cline{5-7} 
                      & \multicolumn{1}{c|}{}                                           &                          &                          & \multicolumn{1}{c|}{$R_{row}=6$ $R_{col}=1$} & 0.0224s & 0.4372s    \\ \cline{5-7} 
                      & \multicolumn{1}{c|}{}                                           &                          &                          & \multicolumn{1}{c|}{$R_{row}=8$ $R_{col}=1$} & 0.0300s & 0.4448s    \\ \hline
\end{tabular}}
\caption{Average computation time for continuous reconstruction on KITTI Sequence 0.}
\label{tab:computation time continuous reconstruction}
\end{table}

Here we briefly describe the computation time of all modules, including upsampling, SPHARM coefficients extraction, and reconstruction, for different reconstruction resolutions tested on the KITTI dataset. The experiment runs on an Intel® Core™ i7-10875H CPU. From TABLE \ref{tab:computation time 1:1}, it can be seen that for 1:1 task, our compression time is slower than the PCL library, although our reconstruction time is much faster. TABLE \ref{tab:computation time continuous reconstruction} shows the run time result of each module by using our default outdoor parameters. Upsampling takes up most of the time since it needs to query the intersection point of each ray to a mesh. 
The computation time of reconstruction increases linearly with respect to the resolution.


\section{Conclusions}
This paper presents CURL, a novel representation pipeline to simultaneously compress and densify LiDAR 3D point clouds. It develops the efficient 2D polar-parametrized dense meshing, the mask-refined upsampling and the adaptive encoding of the spherical harmonic functions for continuous reconstruction. Demonstrated and evaluated on four public datasets, CURL is capable of producing reliable dense reconstructions, e.g., in large-scale outdoor environments, while reducing storage spaces simultaneously. Therefore, it can be found useful in many robotic applications that requires a good balance between fidelity and size of 3D point cloud maps, such as teleoperation, mapping, path planning and navigation using point cloud maps.
Future work will focus on designing dynamic partition of patches, which can further deliver compression and improve reconstruction accuracy.

\section*{Acknowledgments}
This work was supported in part by EU H2020 Programme under DeepField project (grant ID 857339) and SOLITUDE project.

\bibliographystyle{plainnat}
\bibliography{references}

\end{document}